\documentclass[twoside,espcrc2]{elsarticle}
\usepackage{lineno,hyperref}
\modulolinenumbers[5]
\usepackage[margin=3cm]{geometry}
\usepackage{subfigure}
\usepackage{multirow}
\usepackage{bm}
\usepackage{verbatim}
\biboptions{numbers,sort&compress}
\usepackage{color}

\usepackage{array}

\usepackage[table,xcdraw]{xcolor}
\usepackage{booktabs}
\usepackage[numbers]{natbib}
\usepackage{graphicx} 
\usepackage{amssymb}
\usepackage{amsmath}
\usepackage{mathabx}
\usepackage{soul}

\usepackage{hhline}

\usepackage{epstopdf}

\usepackage{enumitem}
\usepackage[font=small,skip=2pt]{caption}
\graphicspath{ {Figures/}}

\newcolumntype{L}[1]{>{\raggedright\let\newline\\\arraybackslash\hspace{0pt}}m{#1}}
\newcolumntype{C}[1]{>{\centering\let\newline\\\arraybackslash\hspace{0pt}}m{#1}}
\newcolumntype{R}[1]{>{\raggedleft\let\newline\\\arraybackslash\hspace{0pt}}m{#1}}

\usepackage{ifthen}
\makeatletter
\renewcommand\@cite[2]{%
Ref.~#1\ifthenelse{\boolean{@tempswa}}
{, \nolinebreak[3] #2}{}
}
\renewcommand\@biblabel[1]{#1.}
\makeatother

\usepackage[textwidth=3.7cm]{todonotes}

\newcounter{cntcomment}

\newcommand{\mr}[1]{\mathrm{#1}}
\newcommand{\tx}[1]{\textrm{#1}}

\newcommand{\vold}[1]{$#1\!\times\!#1\!\times\!#1$}

\renewcommand{\vec}[1]{\mathbf{#1}}

\newcommand{\xx}{\vec x}
\newcommand{\yy}{\vec y}
\newcommand{\bb}{\vec b}
\newcommand{\WW}{\vec w}

\newcommand{\vt}{\boldsymbol{\theta}}

\makeatletter
\def\ps@pprintTitle{%
 \let\@oddhead\@empty
 \let\@evenhead\@empty
 \def\@oddfoot{}%
 \let\@evenfoot\@oddfoot}
\makeatother

\begin{document}

\begin{frontmatter}



\title{Deep CNN ensembles and suggestive annotations for infant brain MRI segmentation}


\author[LIVIA]{*Jose Dolz}
\author[LIVIA]{Christian Desrosiers}
\author[NC]{Li Wang}
\author[XIDIAN]{Jing Yuan}
\author[NC,KU]{*Dinggang Shen}
\author[LIVIA]{*Ismail Ben Ayed}

\address[LIVIA]{Ecole de Technologie Superieure (ETS), University of Quebec, Montreal, QC, Canada}
\address[XIDIAN]{Xidian University, School of Mathematics and Statistics, Xi'an, China}
\address[NC]{Department of Radiology and BRIC, University of North Carolina at Chapel Hill, NC 27599, USA}
\address[KU]{Department of Brain and Cognitive Engineering, Korea University, Seoul 02841, Republic of Korea}

\cortext[mycorrespondingauthor]{Corresponding authors: jose.dolz@livia.etsmtl.ca, dinggang$\_$shen@med.unc.edu, ismail.benayed@etsmtl.ca}
%
%
%

\begin{abstract}

Precise 3D segmentation of infant brain tissues is an essential step towards comprehensive volumetric studies and quantitative analysis of early brain developement. However, computing such segmentations is very challenging, especially for 6-month infant brain, due to the poor image quality, among other difficulties inherent to infant brain MRI, e.g., the isointense contrast between white and gray matter and the severe partial volume effect due to small brain sizes. 

This study investigates the problem with an ensemble of {\em semi-dense} fully convolutional neural networks (CNNs), which employs T1-weighted and T2-weighted MR images as input. We demonstrate that {\em the ensemble agreement is highly correlated with the segmentation errors}. Therefore, our method provides measures that can guide local user corrections. To the best of our knowledge, this work is the first ensemble of 3D CNNs for suggesting annotations within images. Furthermore, inspired by the very recent success of {\em dense} networks \cite{huang2016densely}, we propose a novel architecture, \emph{SemiDenseNet}, which connects all convolutional layers directly to the end of the network. Our architecture allows the efficient propagation of gradients during training, while limiting the number of parameters, requiring one order of magnitude less parameters than popular medical image segmentation networks such as 3D U-Net \cite{cciccek20163d}. Another contribution of our work is the study of the impact that early or late fusions of multiple image modalities might have on the performances of deep architectures. We report evaluations of our method on the public data of the MICCAI iSEG-2017 Challenge on 6-month infant brain MRI segmentation, and show very competitive results among 21 teams, ranking first or second in most metrics.
 
\end{abstract}


\end{frontmatter}


\section{Introduction}

Precise segmentation of infant brain MRI into white matter (WM), gray matter (GM) and cerebrospinal fluid (CSF) is essential to study early brain development. Throughout this period, the postnatal human brain shows its most dynamic phase of development, with a rapid growth of tissues and the formation of key cognitive and motor functions \cite{paus2001maturation}. Infant brain segmentation is also important to detect brain abnormalities occurring shortly after birth, such as hypoxic ischemic encephalopathy, hydrocephalus or congenital malformations, enabling the prediction of neuro-developmental outcomes. 

\begin{figure}[ht!]
     \begin{center}
     \mbox{
        \includegraphics[height=0.30\linewidth]{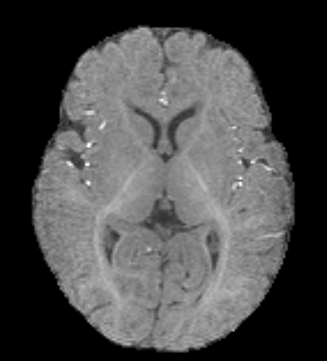}
        \hspace{-2.5 mm}
        \includegraphics[height=0.30\linewidth]{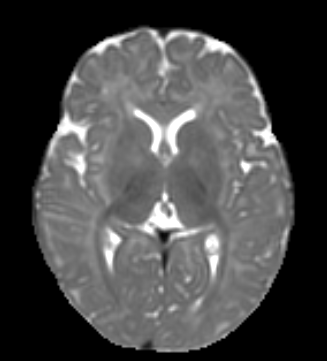}
        \hspace{-2.5 mm}
        \includegraphics[height=0.30\linewidth]{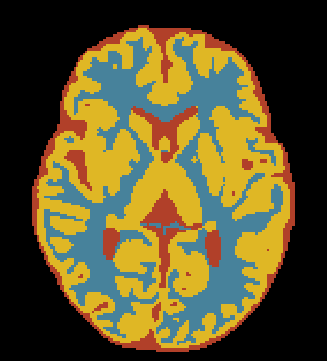}
        }
\caption{Example of data from a training subject.6-month infant brain images from a mid-axial T1w slice (\textit{left}), the corresponding T2w slice (\textit{middle}), and the ground-truth labels (\textit{right}).}
\label{fig:isointense}
\end{center}
\end{figure}

Magnetic resonance imaging (MRI) is commonly used for infant brains because it provides a safe and non-invasive way of examining cross-sectional views of the brain in multiple contrasts. Brain MRI in the first two years can be divided in three distinct phases: infantile ($<$ 6 months), isointense (6-12 months) and early adult-like phase ($>$12 months). Images in the isointense phase show patterns of isointense contrast between white and gray matter (e.g., see Fig. \ref{fig:isointense}), which may vary across brain regions due to nonlinear brain development \cite{paus2001maturation}. These patterns, along with various factors, for instance, limited acquisition time, increased noise, motion artifacts, severe partial volume effect due to smaller brain sizes and ongoing white matter myelination in infant brain images, make automatic segmentation of isointense infant brain MRI a challenging task. 

\subsection{Related work}
\label{ssec:relatedWork}

A popular approach for automatic segmentation uses atlases to model the anatomical variability of target structures \cite{prastawa2005automatic,weisenfeld2006segmentation,song2007clinical,weisenfeld2009automatic,shi2010neonatal,shi2010construction,shi2011infant,melbourne2012neobrains12,cardoso2013adapt,wang2014segmentation}. In such approach, an atlas (or multiple atlases) is first registered to a target image and then used to propagate manual labels to this image. When several atlases are considered, labels from individual atlases can be combined into a final segmentation via a label fusion strategy \cite{weisenfeld2006segmentation,weisenfeld2009automatic,wang2012atlas} such as the STAPLE (Simultaneous Truth and Performance Level Estimation) algorithm \cite{warfield2004simultaneous}. Atlas-based methods have been widely used in a breadth of segmentation problems, e.g.,
parcellation of brain MRI into subcortical structures \cite{dolz2015segmentation}. Although these methods provide state-of-the-art performance in many applications, they are usually sensitive to the registration process, and may fail if the image has a low contrast or the target structure has a large variability. This is particularly problematic in the case of infant brain segmentation, due to isointense contrasts and the high spatial variability of the infant population. 

To overcome the limitations of atlas-based methods, parametric \cite{prastawa2005automatic,xue2007automatic,ledig2012neonatal,wu2012automatic,cardoso2013adapt} or deformable \cite{wang2011automatic} models can be used in a refinement step. Parametric models typically state segmentation as the optimization of an energy function, which integrates pixel/voxel probabilities from the atlas with priors restricting the shape or pairwise interactions of brain tissues \cite{xue2007automatic,ledig2012neonatal,melbourne2012neobrains12,cardoso2013adapt}. 
Such models often need a large number of annotated images, which are rarely available in practice. Deformable models refine atlas-produced contours in an iterative manner so as to align better with image edges. However, these models are typically structure-specific and difficult to extend to multi-tissue segmentation. 


Recently, deep learning methods based on convolutional neural networks (CNNs) have demonstrated outstanding performances in a wide range of computer vision and image analysis applications. In particular, CNNs have achieved state-of-the-art results for various problems \cite{long2015fully,kamnitsas2017efficient,DolzNeuro2017,Fechter_Esophagus}, including the segmentation of infant brain MRI \cite{moeskops2016automatic,zhang2015deep,nie2016fully}. For instance, Moeskops et al. \cite{moeskops2016automatic} presented a multi-scale 2D CNN architecture, which yielded accurate segmentations and spatial consistency using a single image modality (i.e., T2w MRI). To acquire multi-scale information, they considered patches and convolution kernels of multiple sizes. Independent paths were used for patches of different sizes, and the features of these paths were combined at the end of the network. Several recent studies investigated architectures based on multiple modalities as input, in order to overcome the extremely low contrast between WM and GM tissues. For example, Zhang et al. \cite{zhang2015deep} proposed a deep CNN combining MR-T1, T2 and fractional anisotropy (FA) images. Similarly, a fully convolutional neural network (FCNN) was proposed in \cite{nie2016fully} for segmenting isointense phase brain MR images. As further explained in Section \ref{sec:methods}, a FCNN is a special type of CNN that generates dense pixel predictions. Instead of simply stacking the three modalities at the network input, the network in \cite{nie2016fully} processes each modality within an independent path. The final segmentation is obtained by fusing the ensuing paths. These approaches have some important drawbacks. 

First, the architectures in \cite{moeskops2016automatic} and \cite{zhang2015deep} use sliding windows, each defining a region that is processed independently of the other windows. Such a strategy is not efficient because of the many redundant convolution and pooling operations. Furthermore, processing these regions independently yields non-structured predictions, which affects segmentation accuracy. Second, these networks use 2D patches as input. This does not account for the anatomic context in directions orthogonal to the 2D plane. As first shown in \cite{kamnitsas2017efficient}, and later in \cite{DolzNeuro2017}, in different contexts of brain structure segmentation, considering 3D data directly, instead of slice-by-slice, can improve segmentation accuracy. Table \ref{table:table_RV_seg} provides a brief summary of the existing methods for infant brain tissue segmentation. For a detailed review of the methods proposed to address this task, we refer the reader to the recent work of Makropoulos et al. \cite{makropoulos2017review}  

\begin{table*}[h!]
\begin{center}
\begin{scriptsize}
\centering
\renewcommand{\arraystretch}{1.25}
\caption{A brief summary of existing methods for infant brain tissue segmentation.}
\label{table:table_RV_seg}
\begin{tabular}{lC{38mm}C{15mm}C{15mm}C{15mm}}
\toprule
\textbf{Method} & \textbf{Technique} & \textbf{Modality} & \textbf{Stage}** & \textbf{Time}
 \\
 \midrule\midrule
Prastawa et al., 2005 \cite{prastawa2005automatic} & Parametric (Graph clustering) & T1, T2 & IF & -- \\  
Weisenfeld et al., 2006 \cite{weisenfeld2006segmentation} & Atlas + Bayesian Classifier & T1, T2 & IF &  --\\ 
Nishida et al., 2006 \cite{nishida2006detailed} & -- & T1 & IF & --\\
Xue et al., 2007 \cite{xue2007automatic} & Parametric (EM-MRF) & T2 & IS & -- \\
Song et al., 2007 \cite{song2007clinical} & -- & T2 & IF & --\\
Anbeek et al., 2008 \cite{anbeek2008probabilistic} & K-Nearest Neighbours& T2,IR & IF & --\\
Weisenfeld and Warfield, 2009 \cite{weisenfeld2009automatic} & Multi-atlas & T1,T2 & IF & --\\
Shi et al., 2010 \cite{shi2010construction} & Multi-atlas & T2 & IF & -- \\
Wang et al., 2011 \cite{wang2011automatic} & Level-sets & T2 & IF & --\\
Gui et al., 2012 \cite{gui2012morphology} & -- & T1,T2 & IS & 60-75 min\\
Ledig et al., 2012 \cite{ledig2012neonatal} & Parametric (MRF) & T2 & IS & --\\
Makropoulos et al.,2012 \cite{makropoulos2012automatic} & Multi-atlas & T2 &  IS & 95 min\\
Melbourne et al., 2012 \cite{melbourne2012neobrains12} & Parametric (EM-MRF)  & T2 & IS & 15 min\\
Wang et al., 2012 \cite{wang2012atlas} & Multi-atlas & T1, T2 & IS & 7 min\\
Wu et Avants, 2012 \cite{wu2012automatic} & Parametric (MAP-EM) & T1, T2 & IS & 80-100 min\\
Cardoso et al., 2013 \cite{cardoso2013adapt} & Parametric (EM-MRF) & T1 & IS  & --\\
He and Parikh, 2013 \cite{he2013automated} & -- & T2, PD & IF & -- \\
Wang et al., 2014 \cite{wang2014integration} & Multi-atlas & T1,T2,FA & IF/IS/EA & --\\
Wang et al., 2014 \cite{wang2014segmentation} & Multi-atlas & T2 & IF & -- \\
Li et al., 2015 \cite{wang2015links} & Random Forests& T1,T2,FA& IS& 5 min\\
Moeskops et al., 2015 \cite{moeskops2015automatic} & SVM & T2 & IS & -- \\
Zhang et al., 2015 \cite{zhang2015deep} & 2D CNN & T1,T2,FA & IS & 1 min/slice\\
Moeskops et al., 2016 \cite{moeskops2016automatic} & 2D CNN & T2 & IS & -- \\
Nie et al., 2016 \cite{nie2016fully} & 2D CNN & T1,T2,FA & IS & -- \\
\bottomrule
\multicolumn{4}{l}{**Infantile (IF) / Isointense (IS) / Early-adult like (EA)}
\end{tabular}
\end{scriptsize}
\end{center}
\end{table*}

While fully-automatic, learning-based medical image segmentation methods have improved substantially over the last years, the performances in many applications are still insufficient for practical use, more so when the amount of training data is very limited, as is typically the case in medical applications. For instance, manual annotations of brain MRI (i.e., assigning a label to each voxel) is a highly complex and time-consuming process, which requires extensive expertise. This is particularly the case of infant brain MRIs. Therefore, both active and semi/weakly supervised learning frameworks are currently attracting substantial interest in medical image analysis \cite{yang2017suggestive,rajchl2017deepcut,bai2017semi}. For instance, in the recent study in \cite{yang2017suggestive}, Yang et al. proposed an active learning framework, and showed its potential in the context of segmenting glands in histology images and lymph nodes in ultrasound. The purpose of \cite{yang2017suggestive} was to design algorithms that suggest a small set of images to annotate, which lead to the highest possible performance improvement when adding these new annotations to the training set. The framework in \cite{yang2017suggestive} uses an ensemble of deep CNNs to compute an agreement score for candidate instances, and suggests representative instances with the highest uncertainty. However, since suggestions are made at the image level, manual annotations of full images are still required. Using ensemble of CNNs, each trained with a different set of images, can further improve robustness by reducing test error due to variance \cite{kamnitsas2017ensembles}.

\subsection{Contributions and outline}
\label{ssec:contributions}

The contributions of this study can be itemized as follows:

\begin{itemize}

\item This work presents the first ensemble of 3D CNNs for suggesting annotations within images. 
An important benefit of an ensemble of predictors is the ability to measure their level of agreement. This is particularly useful for evaluating the reliability of the segmentations at a voxel level and suggesting local corrections in regions where the ensemble is not confident about the prediction. An important finding in our experiments is that {\em prediction uncertainty, measured as the inverse of predictor agreement within the ensemble, is highly correlated with segmentation errors}.

\item Inspired by the recent success of {\em dense} networks \cite{huang2016densely}, we propose a novel architecture called \emph{SemiDenseNet}, which connects all convolutional layers directly to the end of the network. This semi-dense architecture allows the efficient propagation of gradients during training, while limiting the number of trainable parameters. Our network requires one order of magnitude less parameters than popular medical image segmentation networks such as 3D U-Net \cite{cciccek20163d}. Furthermore, by combining the feature maps of intermediate convolutional layers into the first fully-connected layer, our architecture injects multiscale information into the final segmentation. 

\item As in recent approaches for infant brain MRI segmentation \cite{zhang2015deep,nie2016fully}, the proposed network addresses the problem of low contrast by using multiple image modalities as inputs. So far, to the best of our knowledge, there is no clear guideline in the literature as to how to combine multi-modal images in the network. While stacking available modalities into a single-input network works well in some cases \cite{zhang2015deep}, other works have shown the advantage of having independent paths for modalities and combining these paths further in the network \cite{nie2016fully}. Another contribution of our work is an  investigation of the impact that early or late fusions of several modalities might have on the performances. 

\item We report evaluations of our method on the publicly available data of the MICCAI iSEG-2017 Grand Challenge on 6-month infant brain MRI Segmentation\footnote{http://iseg2017.web.unc.edu/rules/results/}. We obtained very competitive results among 21 teams, ranking first and second in most metrics.


\end{itemize}

The remainder of this paper is as follows. In Section \ref{sec:methods}, we present the proposed semi-dense architecture, and detail how an ensemble of networks can be used to suggest annotations. We also describe the evaluation protocol used in our study. Section \ref{sec:results} demonstrates the performance of our method on data from the MICCAI iSEG-2017 Challenge. Finally, in Section \ref{sec:discussion}, we discuss the main contributions and results of this work, and propose some potential extensions.




\section{Methods}\label{sec:methods}


Convolutional neural networks (CNNs) are a special type of artificial neural networks that learn a hierarchy of increasingly complex features by successive convolution, pooling and non-linear activation operations \cite{lecun1998gradient,krizhevsky2012imagenet}. Originally designed for image recognition and classification, CNNs are now commonly used in semantic image segmentation. A naive approach follows a sliding-window strategy, where regions defined by the window are processed independently. As explained before, this technique presents two main drawbacks: reduction of segmentation accuracy and low efficiency. An alternative approach, known as fully CNNs (FCNNs) \cite{long2015fully}, mitigates these limitations by considering the network as a single non-linear convolution, which is trained in an end-to-end fashion. An important advantage of FCNNs, compared to standard CNNs, is that they can be applied to images of arbitrary size. Moreover, because the spatial map of class scores is obtained in a single dense inference step, FCNNs can avoid redundant convolution and pooling operations, making them computationally more efficient. 


The proposed architectures (Fig. \ref{fig:CNN_archit_Early} and \ref{fig:CNN_archit_Late}) are built on top of \textit{DeepMedic} \cite{kamnitsas2017efficient} and 
extend our network presented in \cite{DolzNeuro2017}, which showed state-of-the-art performance on the task of segmenting subcortical brain structures in MRI. Unlike this network, the proposed architectures use multiple image modalities as input. Moreover, while the previous network has skip-forward connections for only a few convolutional layers, these architectures follow a denser connection strategy, where feature maps from all convolutional layers are aggregated before the first fully-connected layer. Another notable difference is the proposed ensemble learning strategy, where multiple 3D CNNs are combined to improve robustness.

\subsection{Semi-dense 3D fully CNN}
\label{ssec:semi-dense}


Recent hardware developments, in particular those related to graphic processing units (GPUs), have increased the amount of memory available during inference. This has led to a rise in CNN architectures based on 3D convolutions \cite{kamnitsas2017efficient,dou20163d,milletari2016v,lu2017automatic}, which have a much larger number of parameters than their 2D counterpart. To fit volumetric data into memory, 3D CNNs typically perform pooling operations that down-sample feature maps across the network. However, this down-sampling strategy can lead to a loss of resolution in the segmentation. In \cite{long2015fully}, this issue is addressed by connecting features maps of corresponding levels in the down-sampling and up-sampling streams. The resolution of the input image is recovered by adding deconvolution (or \emph{transpose} convolution) layers at the end of the network. Unfortunately, this technique may still give coarse-looking segmentations. For instance, thin structures can disappear after pooling, and may not be recovered in the up-sampling path. To avoid this effect, the proposed FCNN architecture preserves resolution by avoiding down-sampling operations entirely. 

The proposed method extends our recent 3D FCNN achitecture \cite{DolzNeuro2017}, which is composed of many convolutional layers, each containing several 3D convolution filters (or \emph{kernels}). Filters in a layer are applied to the output of the previous layer, or to the input volume in the case of the first layer. The result of this operation is known as a feature map. Let $m_l$ denotes the number of convolution kernels in layer $l$ of the network, and $\xx^n_{l-1}$ the 3D array corresponding to the $n$-th input of layer $l$. The $k$-th output feature map of layer $l$ is then given by
\begin{equation}
    \yy^k_l \ = \ f\Big(\sum^{m_{l-1}}_{n=1} \WW^{k,n}_{i} \otimes \xx^n_{l-1} + \bb^{k}_{l}\Big),
\end{equation} 
where $\WW^{k,n}_{i}$ is a filter convolved with each of the previous layers, $\bb^{k}_{l}$ is the bias, $f$ is a non-linear activation function and $\otimes$ denotes the convolution operator. Note that feature maps produced by convolutions are slightly smaller than their input volumes: The size difference along each dimension is equal to the filter size in this dimension minus one voxel. Hence, applying a \vold{3} convolution filter will reduce the input volume by 2 voxels along each dimension. A stride may also be defined for each convolutional layer, representing the displacement of the filter along the three dimensions after each application.

For the activation function, we used the Parametric Rectified Linear Unit (PReLU) \cite{he2015delving} instead of the popular Rectified Linear Unit (ReLU). This function can be formulated as
\begin{equation}
    f(\xx_i) \ = \ \max(0, \xx_i) \, + \, a_i \! \cdot \! \min(0,\xx_i),
\end{equation}
where $\xx_i$ defines the input signal and $f(\xx_i)$ represents the output. Here, $a_i$ is a scaling coefficient that stops the local gradient from becoming zero when $\xx_i$ is negative. In other words, PReLUs prevent saturation as we approach negative values of $\xx_i$. While ReLU employs predefined values for $a_i$ (typically equal to 0), PReLU requires learning this coefficient. Thus, this activation function can adapt the rectifiers to their inputs, improving the network's accuracy at a negligible extra computational cost. 

As in standard CNNs, fully-connected layers are added at the end of the network to encode semantic information. However, to ensure that the network contains only convolutional layers, we use the strategy described in \cite{long2015fully} and \cite{kamnitsas2017efficient}, in which fully-connected layers are converted to a large set of \vold{1} convolutions. Doing this allows the network to retain spatial information and learn the parameters of these layers as in other convolutional layers. Lastly, neurons in the last layer (i.e., the classification layer) are grouped into $m=C$ feature maps, where $C$ denotes the number of classes. The output of the classification layer $L$ is then converted into normalized probability values via a softmax function. The probability score of class $c \in \{1, \ldots, C\}$ is computed as follows:
\begin{equation}
        p_c \ = \ \frac{\exp\big(y^c_L\big)}{\sum^{C}_{c'=1} \exp\big(y^{c'}_L\big)}
\label{eq:SoftMax}
\end{equation}
 
\begin{figure}[ht!]
\centering
\begin{center} 
        \includegraphics[width=1.02\textwidth]{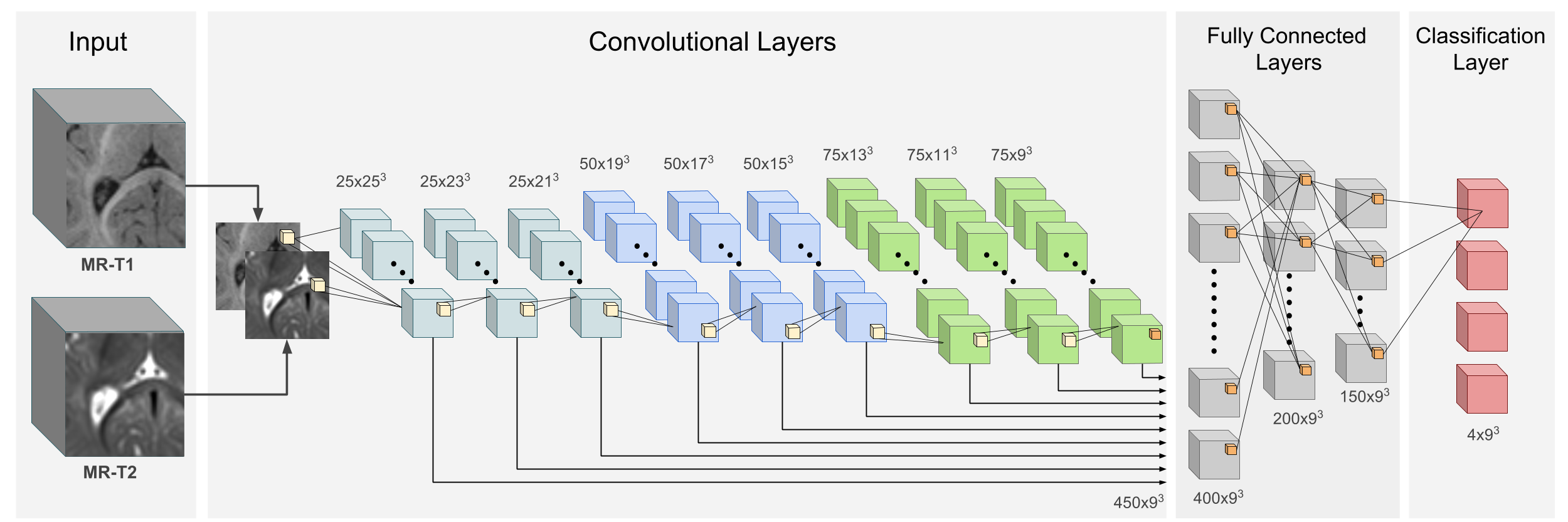}
\caption{Proposed semi-dense architecture using \textit{early fusion} strategy. T1w and T2w MRI inputs are combined before the first convolutional layer and feature maps from every convolutional layer are connected to the first fully-connected layer.}
\label{fig:CNN_archit_Early}
\end{center}        
\end{figure}

\begin{figure}[ht!]
\centering
\begin{center} 
        \includegraphics[width=1.02\textwidth]{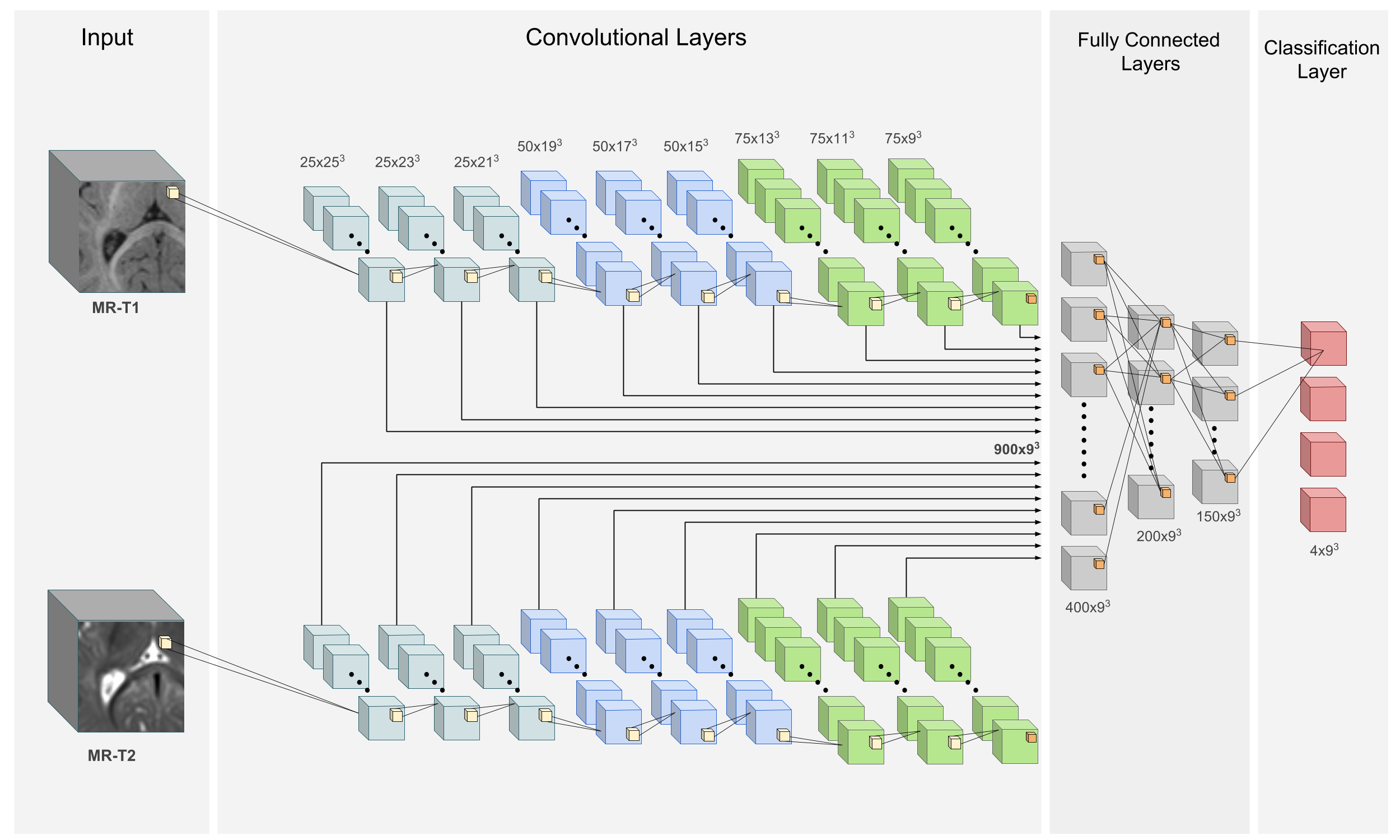}
\caption{Proposed semi-dense architecture using \textit{late fusion} strategy. Features are extracted from T1w and T2w MRI images through independent paths, and are fused before the first fully-connected layer.}
\label{fig:CNN_archit_Late}
\end{center}        
\end{figure}

In addition, we model both local and global context by embedding intermediate-layer outputs in the final prediction. Specifically, we concatenate the output of all convolutional layers into a dense feature map that is fed to the first fully-connected layer. This semi-dense connectivity (Fig. \ref{fig:CNN_archit_Early} and \ref{fig:CNN_archit_Late}), encourages consistency between features extracted at different scales and facilitates the propagation of gradients during training. 



\subsection{Multi-modal input: early versus late fusion}
\label{ssec:multi-source}

The studies in \cite{zhang2015deep,nie2016fully} suggested that multiple input sources provide complementary information, which can help dealing with low contrasts in infant brain MRI. They showed that combining several image sequences in the CNN, in particular T1w, T2w and FA, yielded more accurate segmentations than using these modalities individually. Following this observation, we extend our architecture in \cite{DolzNeuro2017} to accommodate multi-sequence images as input to the network. A common strategy for this task is to merge available modalities at the input of the CNN. This early fusion strategy processes input modalities alongside one another, encouraging their use in all the features of a given layer  \cite{zhang2015deep}. Another approach, referred to as late fusion, employs independent streams for each modality, with features updated separately during training and merged at the end of the network. Considering these strategies, we propose two architectures for our network based on \textit{early} (Fig. \ref{fig:CNN_archit_Early}) or \textit{late} fusion (Fig. \ref{fig:CNN_archit_Late}). 

\subsection{Ensemble learning to suggest local corrections}


Ensemble learning uses multiple classifiers/regressors based on different models or trained with different sets of instances, so as to reduce errors due to variance. The outputs of individual models in the ensemble are combined into a single prediction, for instance, by averaging or majority voting. The justification for this technique lies in the fact that, when multiple and independent decisions are combined, random errors are reduced. 
Thus, ensemble learning promotes better generalization and provides higher prediction accuracy than individual models. Recent studies have demonstrated that averaging the predictions of similar CNNs can lead to an increase in performance. For example, Krizhevsky et al. \cite{krizhevsky2012imagenet} found that, on the ImageNet 2012 classification benchmark, an ensemble of 5 CNNs achieved a top-5 validation error rate of 16.4$\%$, compared to 18.2$\%$ for a single CNN model. By adding another CNN to this ensemble, and making small changes to the network architecture, Zeiler and Fergus \cite{zeiler2014visualizing} were able to decrease this error further to 14.7$\%$.

In our approach, multiple CNNs are generated and combined to obtain an aggregated prediction. Given a training set $\Omega = \{ (\mathcal{X}_t, \yy_t)\}$, $t = 1,\ldots,N$, where $\mathcal{X}_t = (\xx^\mr{T1}_t, \xx^\mr{T2}_t)$ are the T1w and T2w images of training subject $t$ and $\yy_t$ is the corresponding annotated ground truth, we build a set of predictors $\varphi(\xx; \vt_k)$, $k = 1, \ldots, K$, each trained with a randomly selected subset of $\Omega$. This enforces diversity in the predictors, thereby increasing the ensemble's ability to generalize. In this work, we used an ensemble of $K=10$ CNNs and combined their predictions with majority voting. 

An important benefit of an ensemble of predictors is the ability to measure their level of agreement. In \cite{yang2017suggestive}, this measure is used to identify unlabeled images for which the predicted segmentation is uncertain, and to recommend these images to an expert for annotation. In this work, we evaluate the reliability of the segmentation at a voxel level, not at the image level as in \cite{yang2017suggestive}), and suggest local corrections in regions where the ensemble is not confident about the prediction. An important finding in our experiments is that the prediction uncertainty (i.e., the inverse of predictor agreement) is highly correlated with segmentation error.
 
\subsection{Network parameters and implementation details} 

Our CNN is composed of 13 layers in total: 9 convolutional layers in each path, 3 fully-connected layers, and the classification layer. The number of kernels in each convolutional layer, from shallow to deeper, is as follows: 25, 25, 25, 50, 50, 50, 75, 75 and 75. To achieve this depth in a 3D CNN, we employ small kernels of size \vold{3}. Moreover, a unit stride is used for all convolutions to preserve spatial resolution. As in \cite{he2016identity} the activation functions, i.e. PReLU and Batch normalization are employed as 'pre-activation' steps. Thus, each convolutional block is composed by a batch normalization step, followed by a Parametric Rectified Linear Unit (PReLU) activation function and lastly the convolutional filters.

The three fully-connected layers are composed of 400, 200 and 150 hidden units, respectively. Features maps from each convolutional layer are fed into the first fully-connected layer, thereby incorporating multi-scale information in the segmentation. Since the size of feature maps differs from one layer to the next, they are cropped to fit the size of feature maps in the last convolutional layer. Thus, the input to the first fully-connected layer has a size of $\tx{\emph{num. feature maps}} \times  9 \times 9 \times 9$, where the number of feature maps is set to 450 in the early fusion and to 900 in late fusion architectures (Figs. \ref{fig:CNN_archit_Early} and \ref{fig:CNN_archit_Late}, respectively).

Instead of using the whole 3D image as input, we sample $S$ image segments (i.e., sub-volumes) from the image, $\xx_s$, $s=1, \ldots, S$, and feed these segments to the network \cite{kamnitsas2017efficient}. This strategy offers two considerable benefits. First, it reduces the memory requirements of our network, thereby removing the need for spatial pooling. More importantly, it substantially increases the number of training examples without having to perform data augmentation. Network parameters are optimized via the RMSprop optimizer, using cross-entropy as cost function. Let $\vt$ denotes the network parameters (i.e., convolution weights, biases and $a_i$ from the rectifier units), and $y^v_s$ the label of voxel $v$ in the $s$-th image segment. Following the training scheme proposed in \cite{kamnitsas2017efficient}, we define the following cost function:
\begin{equation}
    J(\vt) \ = \ 
    -\frac{1}{S\!\cdot\!V} \sum^{S}_{s=1} \sum^{V}_{v=1} \sum^{C}_{c=1} \delta(y^v_s = c) \cdot \log \, p^v_c(\xx_s),
\end{equation}
where $p^v_c(\xx_s)$ is the output of the network for voxel $v$ and class $c$, when the input segment is $\xx_s$. 
In \cite{kamnitsas2017efficient}, Kamnitsas et al. found that increasing the size of input segments in training leads to a higher 
performance, but this performance increase stops beyond segment sizes of \vold{25}. In their network, using this segment size for training, score
maps at the classification stage were of size \vold{9}. Since our architecture is one layer deeper, and to keep the same score map sizes, we set 
the segment size in our network to \vold{27}. This method was used in \cite{DolzNeuro2017} with very satisfactory results.

The initialization of weights in deep CNNs is usually performed by assigning random normal-distributed values to kernel and bias weights. However, using fixed standard deviations to initialize weights might lead to poor convergence \cite{simonyan2014very}. To overcome this problem, we adopted the strategy proposed in \cite{he2015delving}, and used in \cite{DolzNeuro2017,kamnitsas2017efficient} for segmentation, which allows very deep architectures to converge rapidly. We use a zero-mean Gaussian distribution of standard deviation $\sqrt{2/n_l}$ to initialize the weights in layer $l$, where $n_l$ denotes the number of connections to units within layer $l$. 
 
We set momentum to 0.6 and the initial learning rate to 0.001, with the latter reduced by a factor of 2 after every 5 epochs (starting from epoch 10). 
Instead of employing an adaptive strategy for the learning rate, we used step decay and monitored the evolution of the cost function during training. 
We observed that it followed an exponentially decreasing curve with small increasing/decreasing slopes and, therefore, kept this simple yet effective strategy. 
Our 3D FCNNs were trained for 30 epochs, each consisting of 20 subepochs. At each subepoch, a total of 1000 samples were randomly selected from the training images, and processed in batches of size 20.



The code of the proposed 3D FCNN architecture is implemented in Theano \cite{bergstra2010theano} and is publicly available\footnote{https://www.github.com/josedolz/SemiDenseNet}. Training and testing were performed on a server equipped with an NVIDIA Tesla P100 GPU with 16 GB of RAM memory. Training a single network takes around 30 min per epoch, and around 17 hours in total. The segmentation of a 3D MR scan requires 10 seconds per CNN model, on average. On a single GPU, segmenting a new subject with the ensemble of CNNs takes around 100 seconds.

\subsection{Dataset}

The images were acquired at the UNC-Chapel Hill on a Siemens head-only 3T scanner with a circular polarized head coil, and were randomly chosen from the pilot study of the Baby Connectome Project (BCP)\footnote{http://babyconnectomeproject.org}. During scan, infants were asleep, unsedated and fitted with ear protection, with the head secured in a vacuum-fixation device.

\paragraph{Acquisition parameters}

T1-weighted images were acquired with 144 sagittal slices using the following parameters: TR/TE = 1900/4.38 ms, flip angle = 7$^\circ$, resolution = 1$\times$1$\times$1 mm$^3$. Likewise, T2-weighted images were obtained with 64 axial slices by employing: TR/TE = 7380/119 ms, flip angle = 150$^\circ$, resolution =1.25$\times$1.25$\times$1.95 mm$^3$.

\paragraph{Pre-Processing}

The preprocessing was performed by the iSEG-2017 organizers. Specifically, T2w images were linearly aligned onto their corresponding T1w images. All images were resampled into an isotropic 1$\times$1$\times$1 mm$^3$ resolution. Standard image pre-processing steps were then applied using in-house tools, including skull stripping, intensity inhomogeneity correction, and removal of the cerebellum and brain stem. This pre-processing was performed to eliminate the effects that different image registration and bias correction algorithms might have on infant brain segmentation. 

\paragraph{Ground truth generation}

The manual labels were prepared by the iSEG-2017 organizers. Instead of starting from scratch, an initial automatic segmentation for 6-month subjects \cite{wang2013longitudinally,wang20124d} was generated with the guidance from follow-up 24-month scans with high tissue contrast, using the publicly-available iBEAT tool\footnote{http://www.nitrc.org/projects/ibeat/}. Based on this initial automatic segmentation, manual editing was then performed by an experienced neuro-radiologist, to correct segmentation errors in both T1- and T2-weighted MR images. Geometric defects were also removed with the help of surface rendering, using ITK-SNAP. For example, if a hole/handle was found on the surface, the neuro-radiologist first localized the related slices and then checked the segmentation maps of both T1w and T2w images, in order to determine whether to fill the hole or cut the handle. Using this approach, correcting segmentation of a single subject took about one week.



\subsubsection{Evaluation}
\label{sssec:evaluation}

The MICCAI iSEG-2017 organizers used three metrics to evaluate the accuracy of the competing segmentation methods: Dice Similarity Coefficient (DSC) \cite{dice1945measures}, Modified Hausdorff distance (MHD), where the 95-\textit{th} percentile of all Euclidean distances is employed, and Average Surface Distance (ASD). The first measures the degree of overlap between the segmentation region and ground truth, whereas the other two evaluate boundary distances. 

\paragraph{Dice similarity coefficient (DSC)}

Let $V_\mr{ref}$ and $V_\mr{auto}$ be, respectively, the reference and automatic segmentations of a given tissue class and for a given subject. The DSC can be defined as:
\begin{equation}
\mr{DSC}\big(V_\mr{ref}, V_\mr{auto} \big) \ = \ 
    \frac{2 \mid V_\mr{ref} \cap V_\mr{auto}\mid} {\mid V_\mr{ref}\mid +\mid V_\mr{auto}\mid}
\end{equation}
DSC values are within a $[0,1]$ range, 1 indicating perfect overlap and 0 corresponding to a total mismatch.

\paragraph{Modified Hausdorff distance (MHD)}

Let $P_\mr{ref}$ and $P_\mr{auto}$ denote the sets of voxels within the reference and automatic segmentation boundary, respectively. MHD is given by:
\begin{equation}
\mr{MHD}\big(P_\mr{ref}, P_\mr{auto} \big) \ = \ \max \Big\{ \max_{q \in P_\mr{ref}}d(q,P_\mr{auto}), \max_{q \in P_\mr{auto}}d(q,P_\mr{ref}) \Big\},
\end{equation}
where $d(q,P)$ is the point-to-set distance defined by: $d(q,P)=\min_{p \in P} \| q-p\|$, with $\|.\|$ denoting the Euclidean distance. 

\paragraph{Average surface distance (ASD)}

Using the same notation as the Hausdorff distance above, the ASD corresponds to:
\begin{equation}
    \mr{ASD}\big(P_\mr{ref}, P_\mr{auto} \big) \ = \ 
     \frac{1}{|P_\mr{ref}|} \sum_{p \, \in \, P_\mr{ref}} 
     d(p, P_\mr{auto}),
\end{equation}
where $|.|$ denotes the cardinality of a set. 
In distance-based metrics, smaller values indicate higher proximity between two point sets and, thus, a better segmentation.

\section{Results}\label{sec:results}

Three different methods are evaluated in our experiments. The first, referred to as \textit{EarlyFusion$\_$Single}, is a semi-dense network with early fusion of multi-modal images (Fig. \ref{fig:CNN_archit_Early}). The second, denoted \textit{EarlyFusion$\_$Ensemble}, consists of an ensemble of $10$ early-fusion CNNs, trained with different subjects. Finally, the third method, referred to as \textit{LateFusion$\_$Ensemble}, is an ensemble of 10 semi-dense CNNs, each performing a late fusion of modalities in different paths (Fig. \ref{fig:CNN_archit_Late}) and trained with different subjects. 
For the \textit{EarlyFusion$\_$Single} approach, we used 9 subjects for training and one for validation. In the ensemble methods, for each CNN, 8 different subjects were used for training and 2 subjects for validation.

Table \ref{table:results} reports the results obtained by the three proposed methods and the top 5 among the 21 teams that participated in the iSEG MICCAI Grand Challenge\footnote{http://iseg2017.web.unc.edu/rules/results/}. In this table, mean DSC, MHD and ASD values are given separately for CSF, WM and GM tissues. We first observe that adopting an ensemble learning strategy (i.e., \textit{EarlyFusion\_Ensemble}) results in a small yet noticeable improvement in global performance, with respect to our baseline using a single CNN (i.e., \textit{EarlyFusion$\_$Single}). The results also indicate that fusing features from image modalities in a late stage does not help the segmentation compared to an early fusion strategy. In fact, improvements in the \textit{LateFusion\_Ensemble} approach only occurred for 2 out of 9 combinations of metric and tissue (ASD for CSF and MHD for GM). 
Comparing against other competing approaches, the proposed \textit{EarlyFusion\_Ensemble} method obtained the best score in 5 out of 9 cases. In particular, our network yielded the best DSC values for the three brain tissues, and the best MHD and ASD values for CSF and white matter, respectively. Furthermore, all tested methods obtained the highest accuracy for CSF, and slightly better results for GM than WM. As can be seen in Fig. \ref{fig:isointense}, the edges between GM and WM tissues are weak, resulting in a harder classification for voxels lying on these edges. 

\begin{table}[ht!]
\centering
\caption{Segmentation results from the iSEG-2017 Segmentation challenge for the top-5 ranked methods. The first set of the results correspond to the three proposed approaches, and the second set to approaches presented by the other competing teams in the top-5 (in alphabetical order). Bold fonts indicate the best performances for each metric and structure. For additional details, we refer the reader to the challenge's website.}
\label{table:results}
\begin{small}
\begin{tabular}{lccccccccc}
\toprule
\multirow{2}[3]{*}{\textbf{Method}} & \multicolumn{3}{c}{{CSF}} & \multicolumn{3}{c}{{GM}} & \multicolumn{3}{c}{{WM}} \\
\cmidrule(lr){2-4}\cmidrule(lr){5-7}\cmidrule(lr){8-10}
& {DSC} & {MHD} & {ASD} & \multicolumn{1}{l}{{DSC}} & \multicolumn{1}{l}{{MHD}} & \multicolumn{1}{l}{{ASD}} & \multicolumn{1}{l}{{DSC}} & \multicolumn{1}{l}{{MHD}} & \multicolumn{1}{l}{{ASD}} \\
\midrule\midrule
\textbf{EarlyFusion$\_$Single}      & 0.95  & 9.30  & 0.13 & \textbf{0.92} & 7.13 & 0.35 & \textbf{0.90} & 6.90 & 0.41  \\
\textbf{EarlyFusion$\_$Ensemble} & \textbf{0.96} & \textbf{9.03} & 0.14 & \textbf{0.92} & 6.42  & 0.34  & \textbf{0.90}  & 6.98  & \textbf{0.38}\\
\textbf{LateFusion$\_$Ensemble} & \textbf{0.96} & 9.13   & \textbf{0.12}  &  \textbf{0.92} & 6.06 & 0.34  & \textbf{0.90}  & 7.45 & 0.41\\
\midrule
Bern\_IPMI & \textbf{0.96} & 9.62 & 0.13 & \textbf{0.92} & 6.46 & 0.34 & \textbf{0.90} & 6.78 & 0.40 \\
MSL\_SKKU & \textbf{0.96} & 9.07 & \textbf{0.12} & \textbf{0.92} & \textbf{5.98} & \textbf{0.33} & \textbf{0.90} & \textbf{6.44} & 0.39 \\
nic\_vicorob & 0.95 & 9.18 & 0.14 & 0.91 & 7.65 & 0.37 & 0.89 & 7.15 & 0.43 \\
TU/e IMAG/e & 0.95  & 9.43  & 0.15 & 0.90 & 6.86 & 0.38 & 0.89 & 6.91 & 0.43  \\

\bottomrule
\end{tabular}
\end{small}
\end{table}

\begin{table}[ht!]
\centering
\caption{Segmentation results obtained by the proposed \textit{EarlyFusion$\_$Ensemble} approach on the 13 test subjects of the iSEG Segmentation challenge.}
\label{table:resultsSingle}
\begin{small}
\begin{tabular}{lccccccccc}
\toprule
\multirow{2}[3]{*}{\textbf{Subject ID}} & \multicolumn{3}{c}{{CSF}} & \multicolumn{3}{c}{{GM}} & \multicolumn{3}{c}{{WM}} \\
\cmidrule(lr){2-4}\cmidrule(lr){5-7}\cmidrule(lr){8-10}
& {DSC} & {MHD} & {ASD} & \multicolumn{1}{l}{{DSC}} & \multicolumn{1}{l}{{MHD}} & \multicolumn{1}{l}{{ASD}} & \multicolumn{1}{l}{{DSC}} & \multicolumn{1}{l}{{MHD}} & \multicolumn{1}{l}{{ASD}} \\
\midrule\midrule
\textbf{\#11} & 0.9637 & 7.5498 & 0.1024 & 0.9300 & 4.3589 & 0.2838 &  0.9065 & 8.4853 & 0.3423 \\
\textbf{\#12} & 0.9526 & 6.7823 & 0.1287 & 0.9037 & 6.4807 & 0.3780 & 0.8702 & 6.6333  &  0.4558 \\
\textbf{\#13} & 0.9619 & 10.1980 & 0.1158 & 0.9276 & 6.4031 & 0.3172 & 0.9104 & 9.4868 & 0.3685 \\ 
\textbf{\#14} & 0.9423 & 8.9443  & 0.1565 & 0.9091 & 6.3246  & 0.3767 &  0.8912 & 5.7446 &  0.4203 \\
\textbf{\#15} & 0.9607 & 9.2736  & 0.1064 & 0.9281 & 4.5826 & 0.3159 & 0.9054 & 7.0000 & 0.3728 \\
\textbf{\#16} & 0.9582 & 10.8167 & 0.1179 & 0.9208 & 8.1240  & 0.3237 & 0.9111  & 6.5574 & 0.3733 \\
\textbf{\#17} & 0.9609 & 8.1240 & 0.0405 & 0.9270 & 7.8740 & 0.2828 & 0.9119 & 5.9161 & 0.3277 \\
\textbf{\#18} & 0.9646 & 9.4340 & 0.1040 & 0.9201 & 6.4031 & 0.3151 & 0.9033 & 7.1414 & 0.3735 \\
\textbf{\#19} & 0.9598 & 9.0000 & 0.1111 & 0.9202 & 5.6569 & 0.3198  & 0.9088 & 8.1854 & 0.3830 \\
\textbf{\#20} & 0.9524 & 10.0499 & 0.1352 & 0.9039 & 6.7082 & 0.4207 & 0.8690 & 5.9161 & 0.5073 \\
\textbf{\#21} & 0.9587 & 9.4340 & 0.1132 & 0.9156  & 5.8310 & 0.3343 &  0.8908 & 5.8310 & 0.4275 \\
\textbf{\#22} & 0.9454 & 8.7750 & 0.4491 & 0.9158 & 8.4853 & 0.3951 & 0.8896 & 7.0711 & 0.1248 \\
\textbf{\#23} & 0.9598 & 9.0000 & 0.1087 & 0.9196  & 6.1644 & 0.3339  & 0.8934 & 6.7082 & 0.4080 \\
\bottomrule
\end{tabular}
\end{small}
\end{table}

Considering results for individual test subjects (Table \ref{table:resultsSingle}), the proposed \textit{Early\_Ensemble} approach yields an accurate segmentation in most cases, thus showing its robustness. A lower performance was, however, obtained for a few cases, for example, segmenting the CSF of \textit{subject\_022}, which yielded an ASD of 0.449 mm. Ignoring this result brings the mean ASD for CSF down to 0.11, which is lower than the best ASD value of 0.12 for this tissue. 
In a paired t-test, 
our approach performs at the same level as the best competing method (i.e., MSL\_SKKU), with no significant difference (p $>$ 0.01) observed in any of the test cases. Note that this competing method is also based on deep CNNs. 

\begin{figure}[ht!]
     \mbox{
        \includegraphics[width=0.24\linewidth]{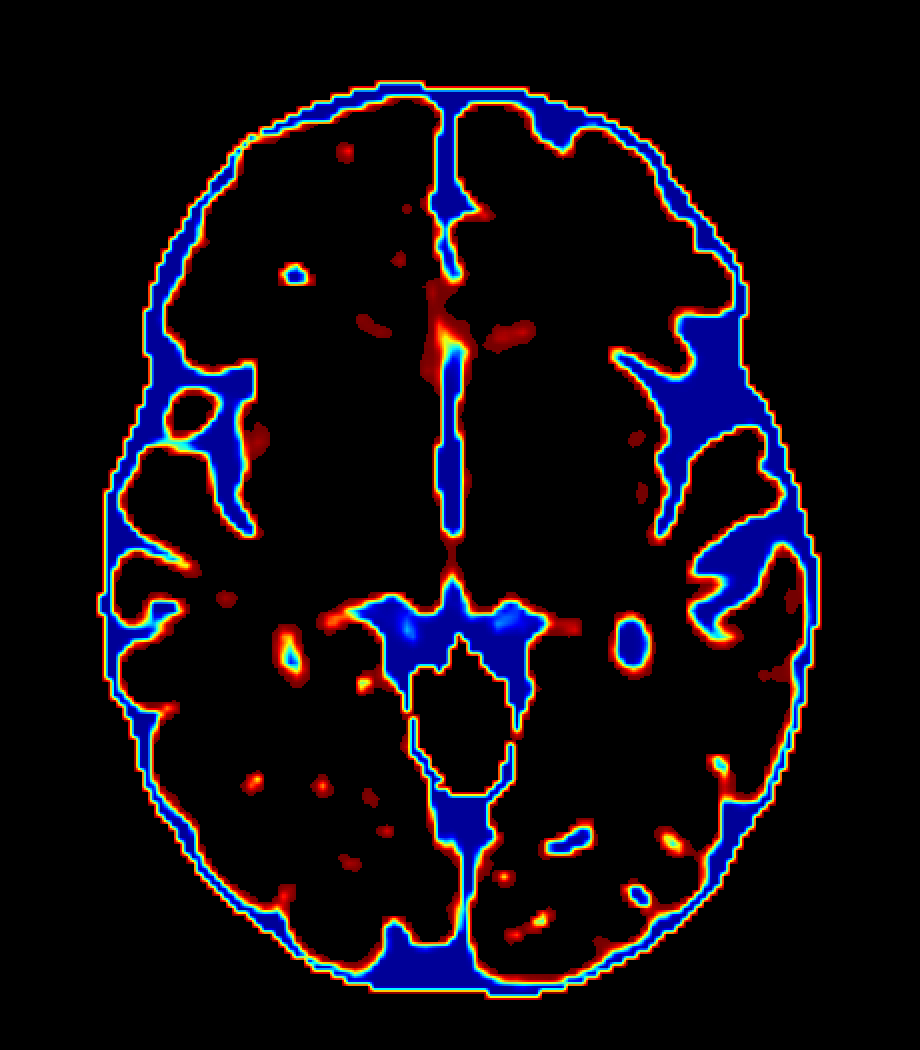}
        \hspace{-2.25 mm}
        \includegraphics[width=0.24\linewidth]{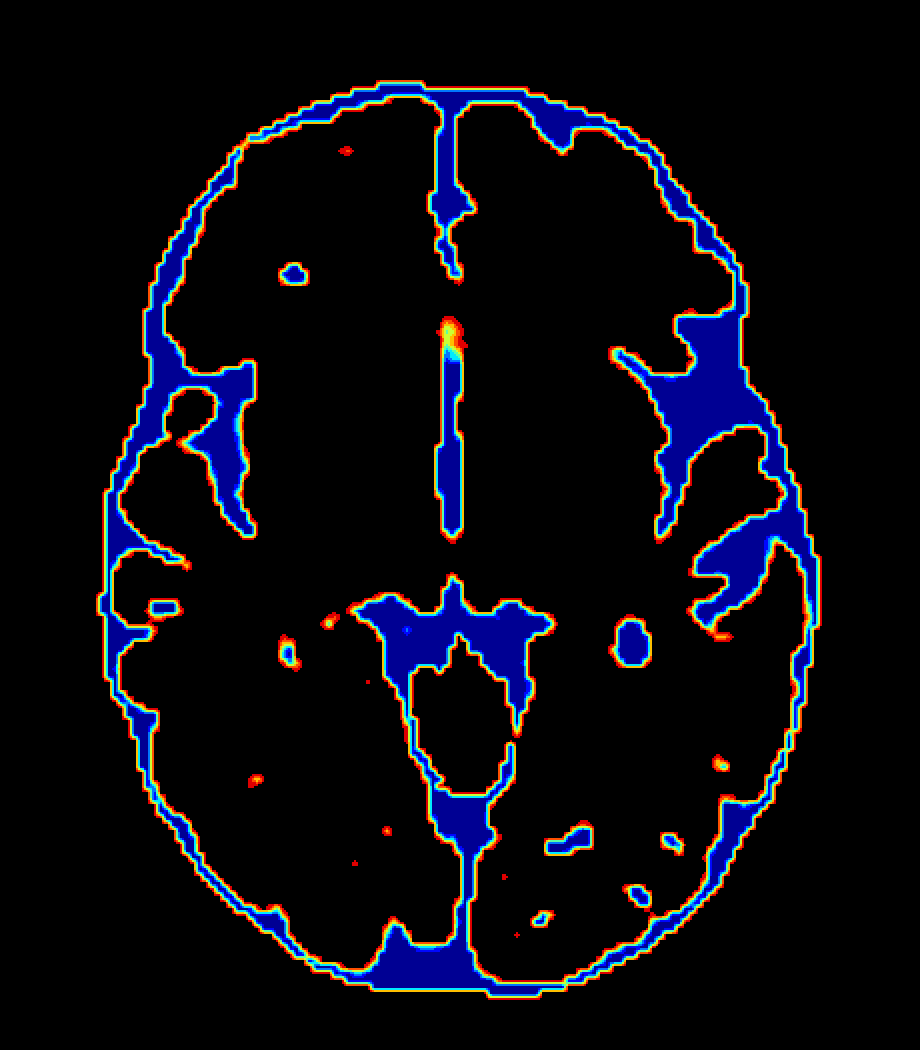}
        \hspace{-.25 mm}
        \includegraphics[width=0.24\linewidth]{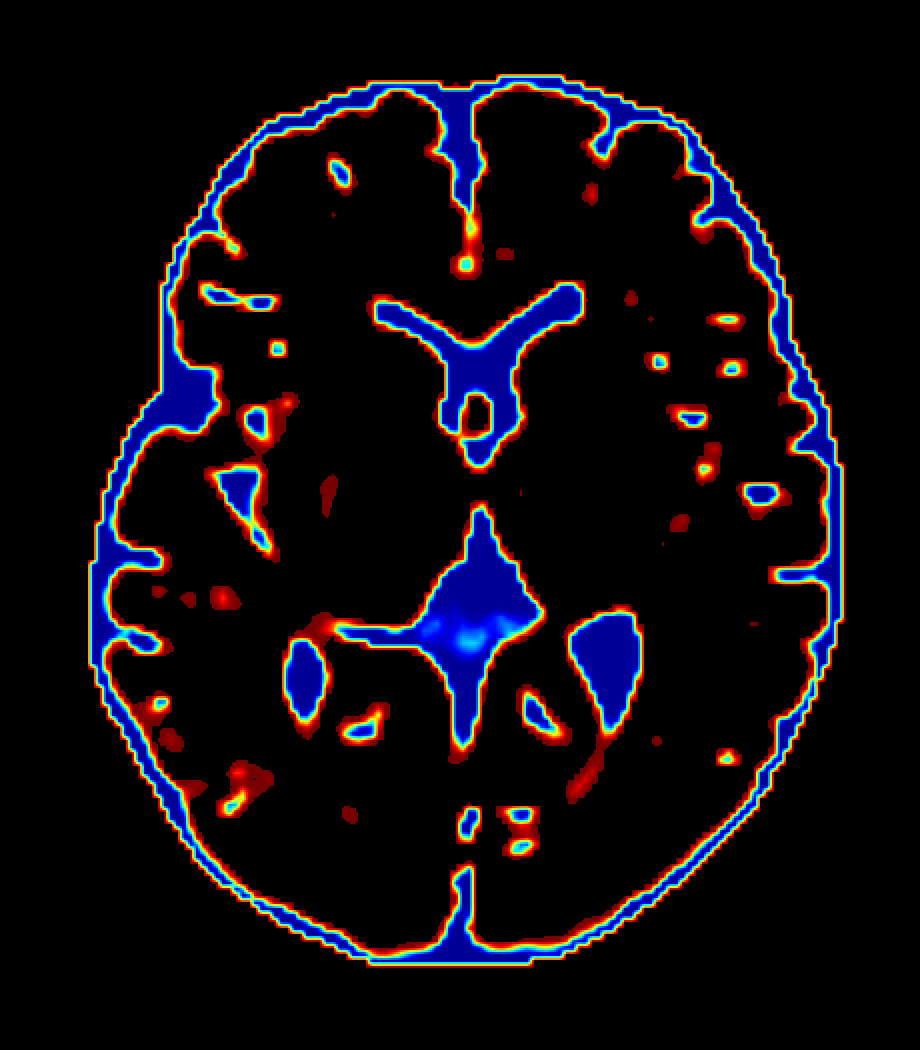}
        \hspace{-2.25 mm}
        \includegraphics[width=0.24\linewidth]{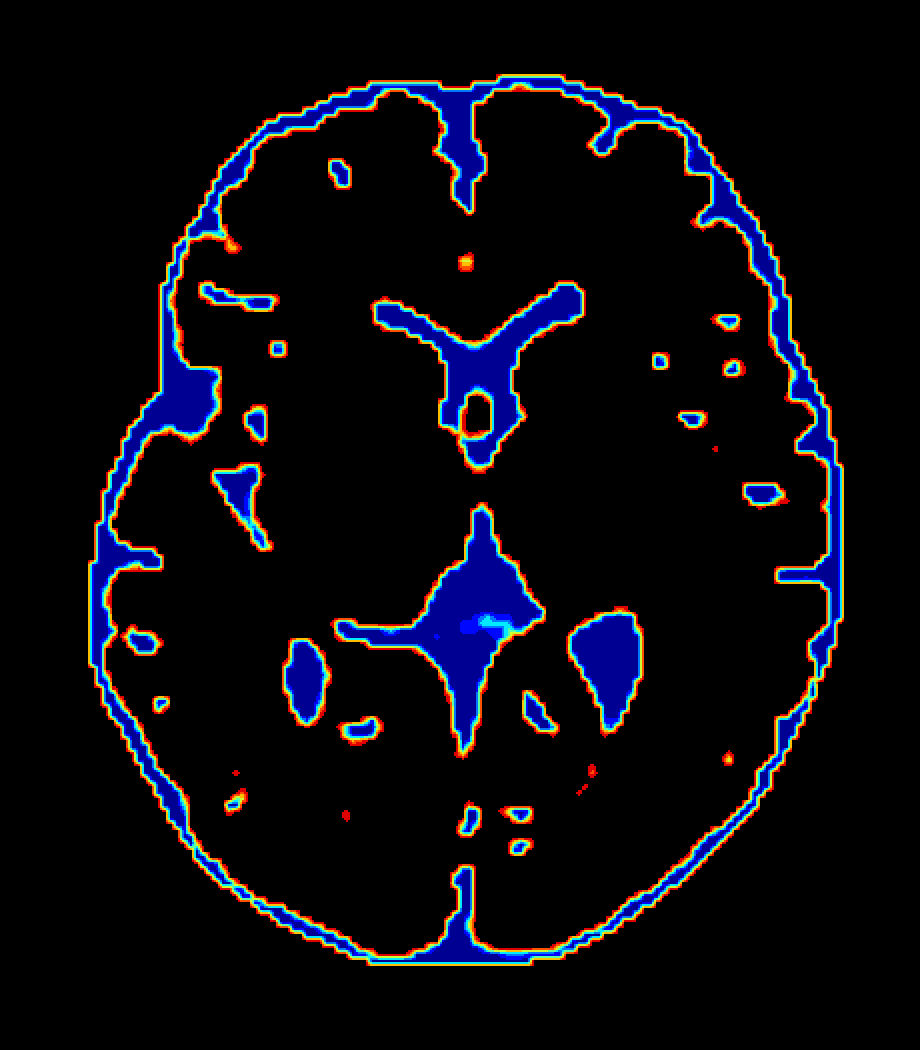}
        }
             
        \vspace{1mm}
        
         \mbox{
        \includegraphics[width=0.24\linewidth]{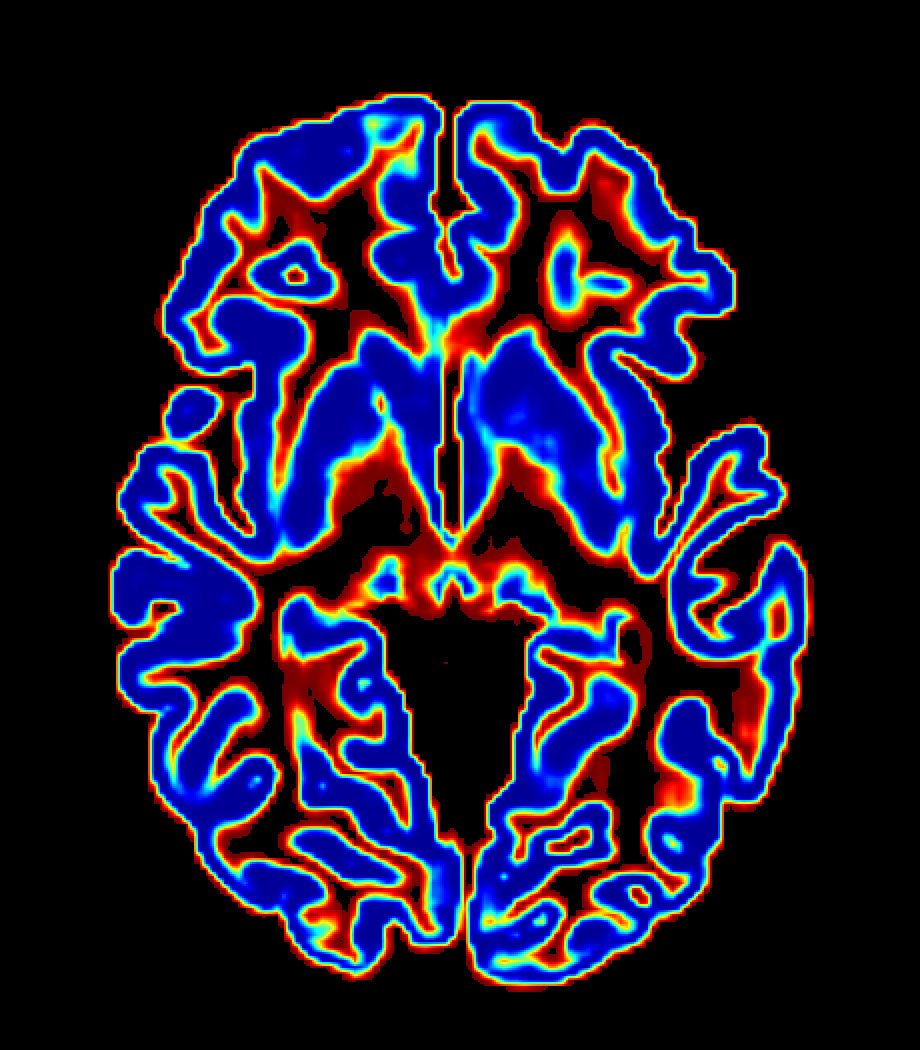}
        \hspace{-2.25 mm}
        \includegraphics[width=0.24\linewidth]{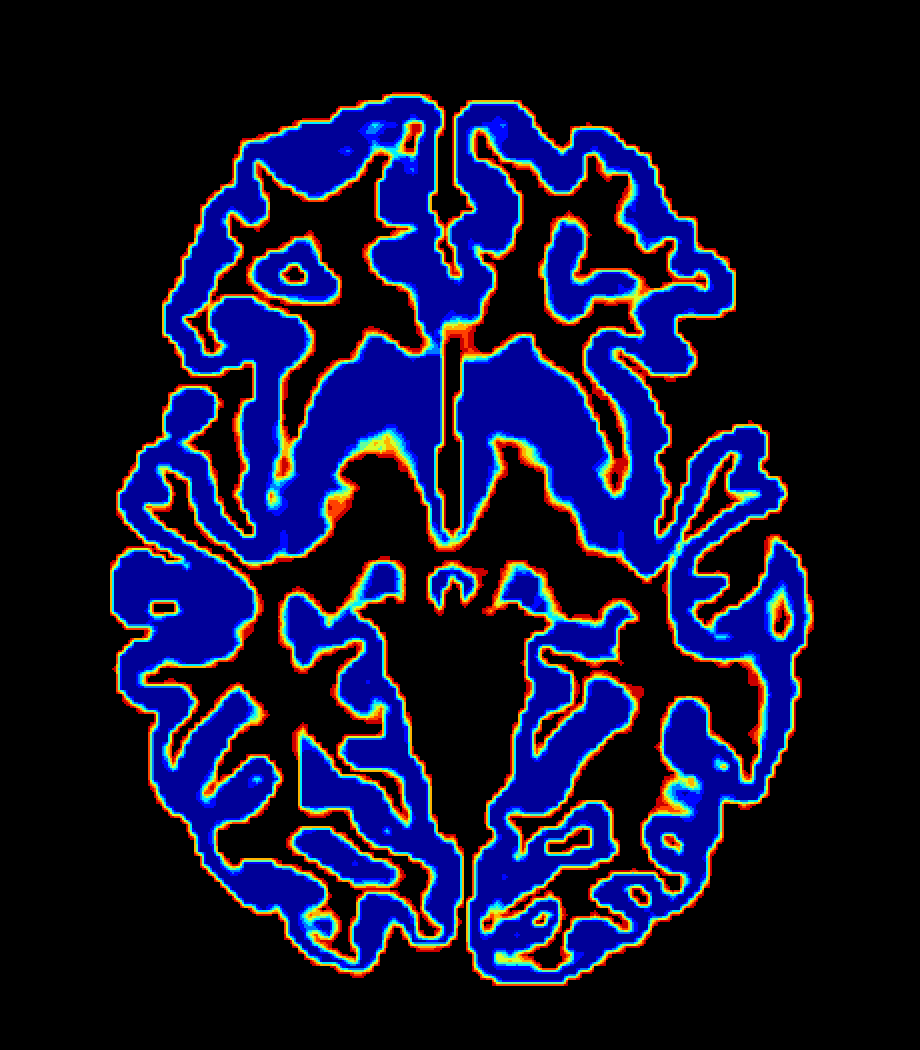}
        \hspace{-.25 mm}
        \includegraphics[width=0.24\linewidth]{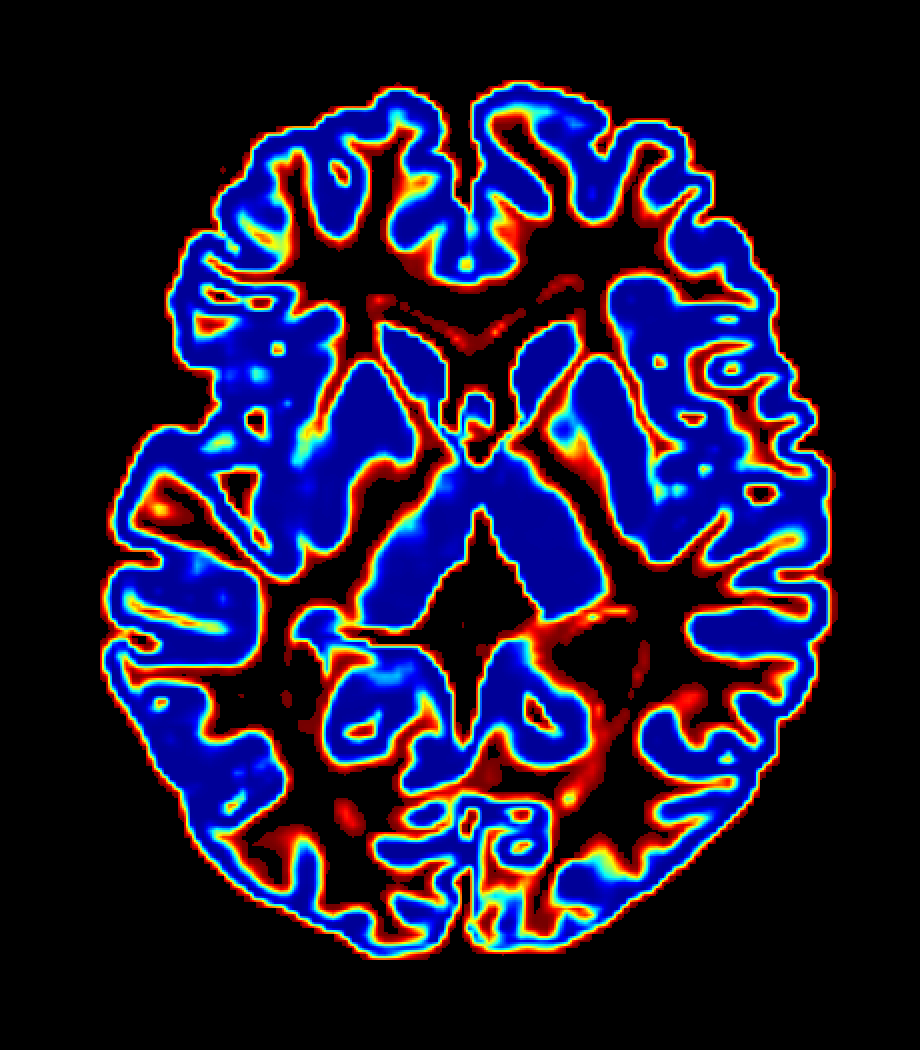}
        \hspace{-2.25 mm}
        \includegraphics[width=0.24\linewidth]{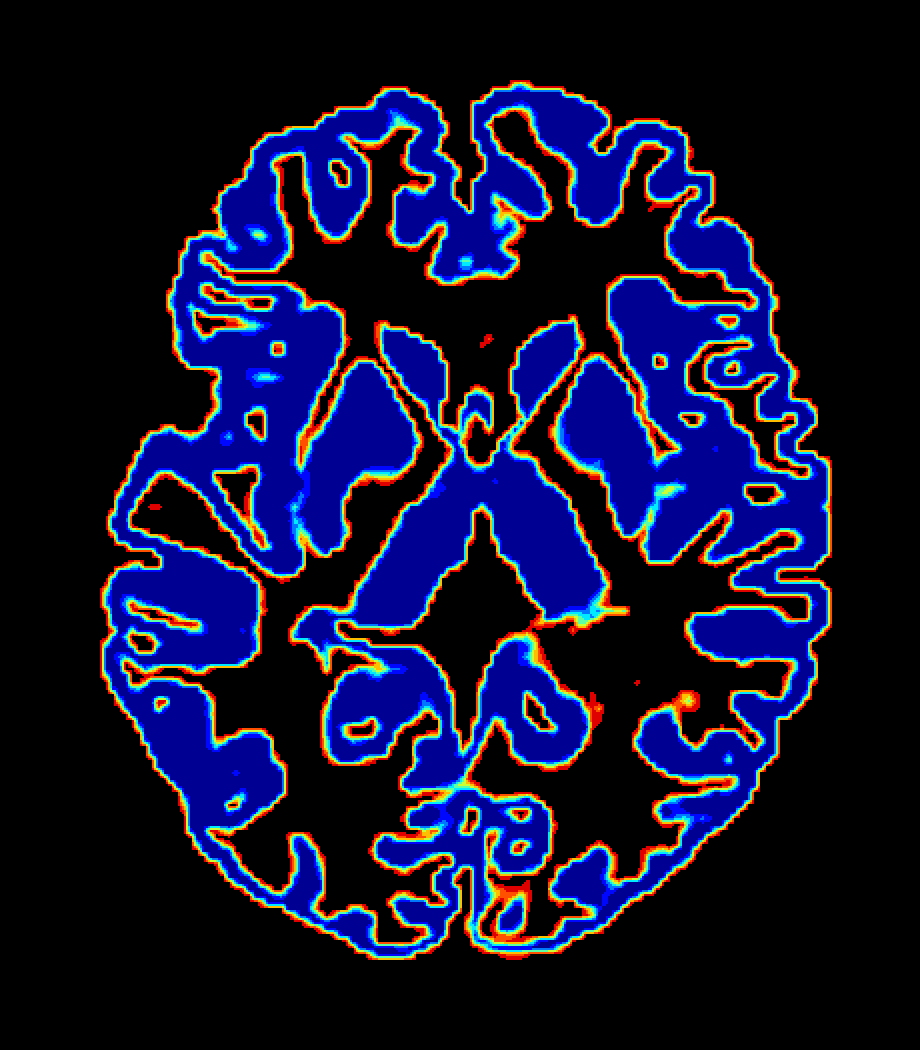}
        }
        
        \vspace{1mm}
        
      \mbox{
        \includegraphics[width=0.24\linewidth]{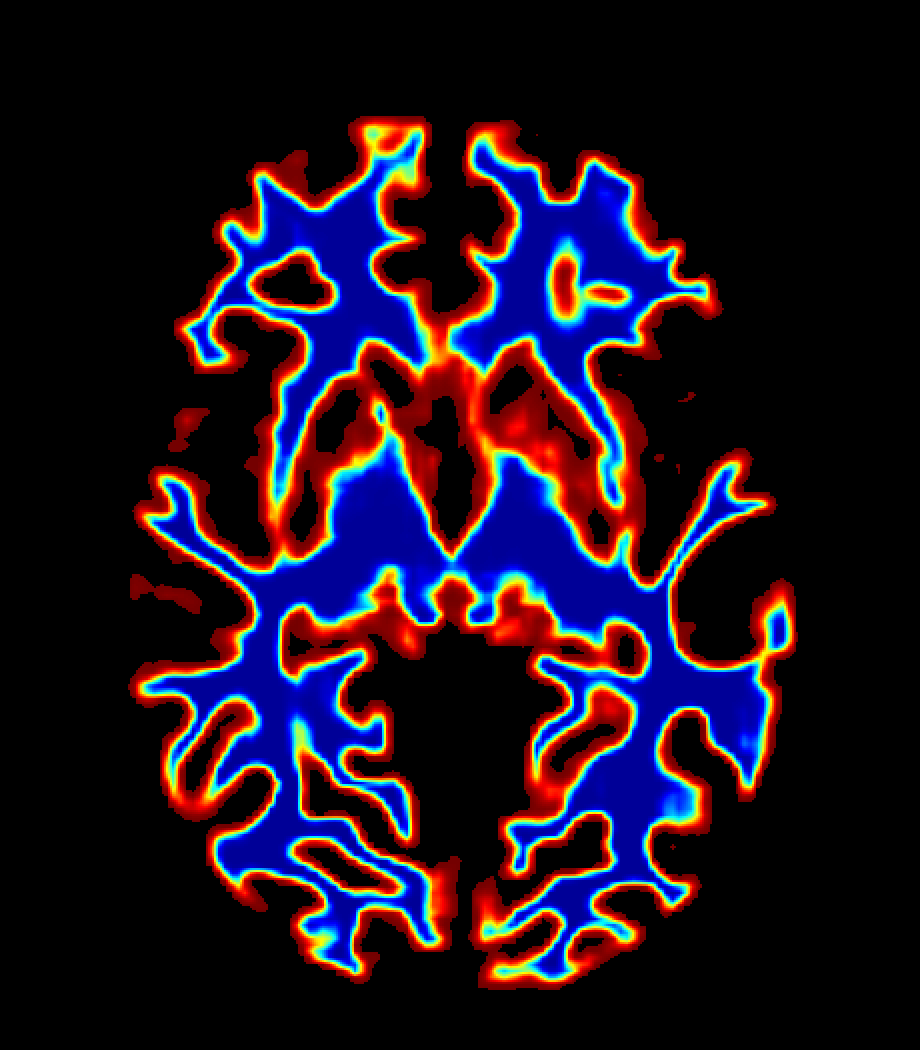}
        \hspace{-2.25 mm}
        \includegraphics[width=0.24\linewidth]{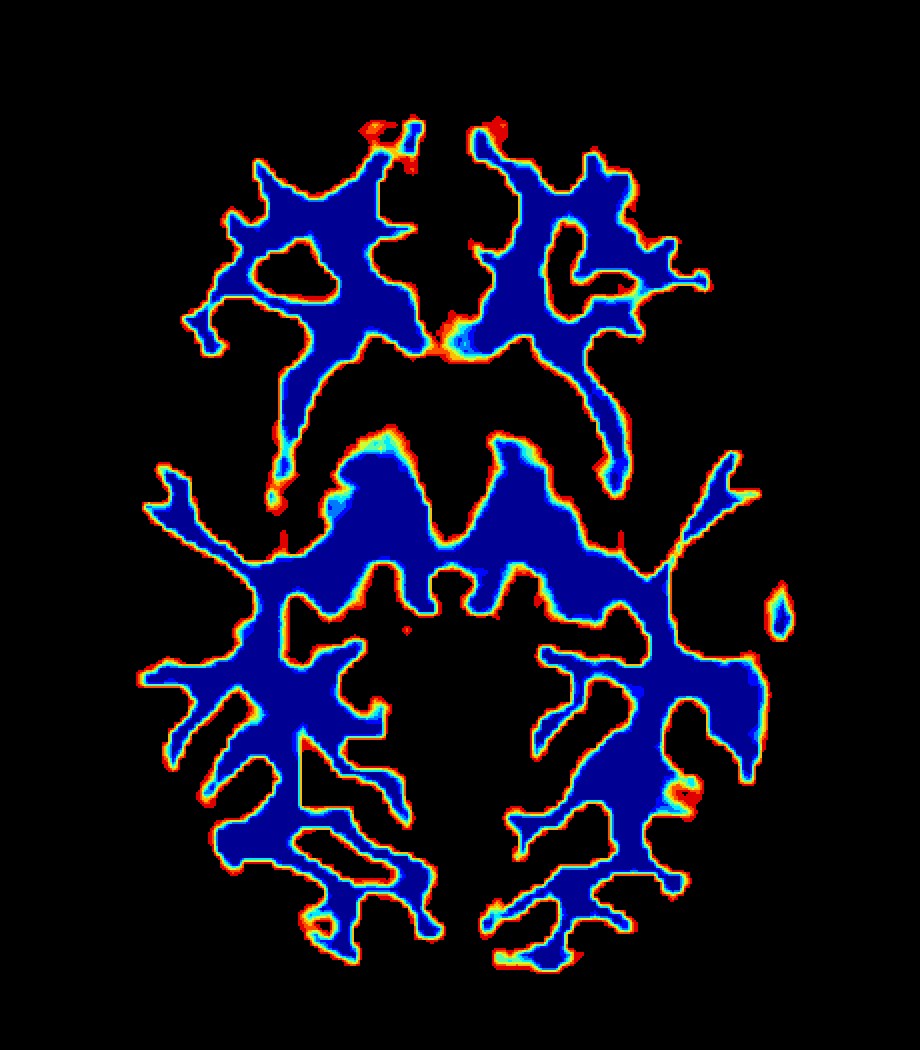}
        \hspace{-.25 mm}
        \includegraphics[width=0.24\linewidth]{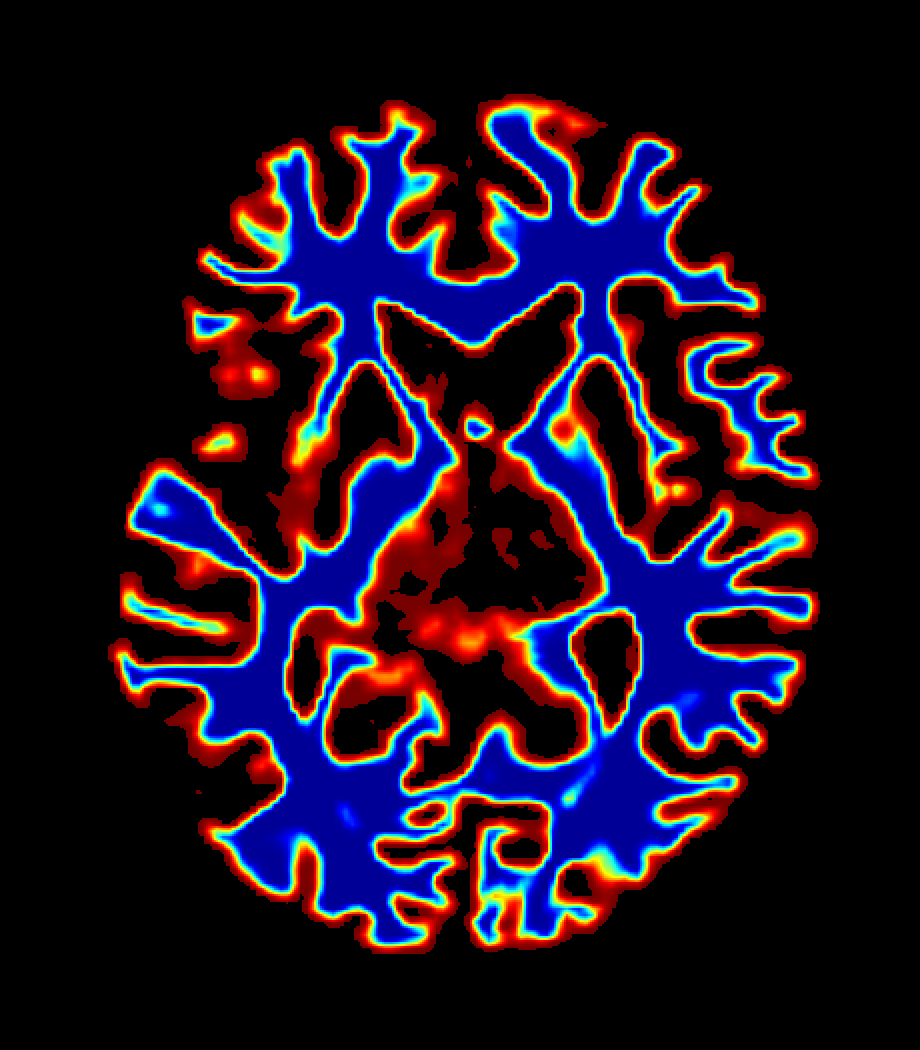}
        \hspace{-2.25 mm}
        \includegraphics[width=0.24\linewidth]{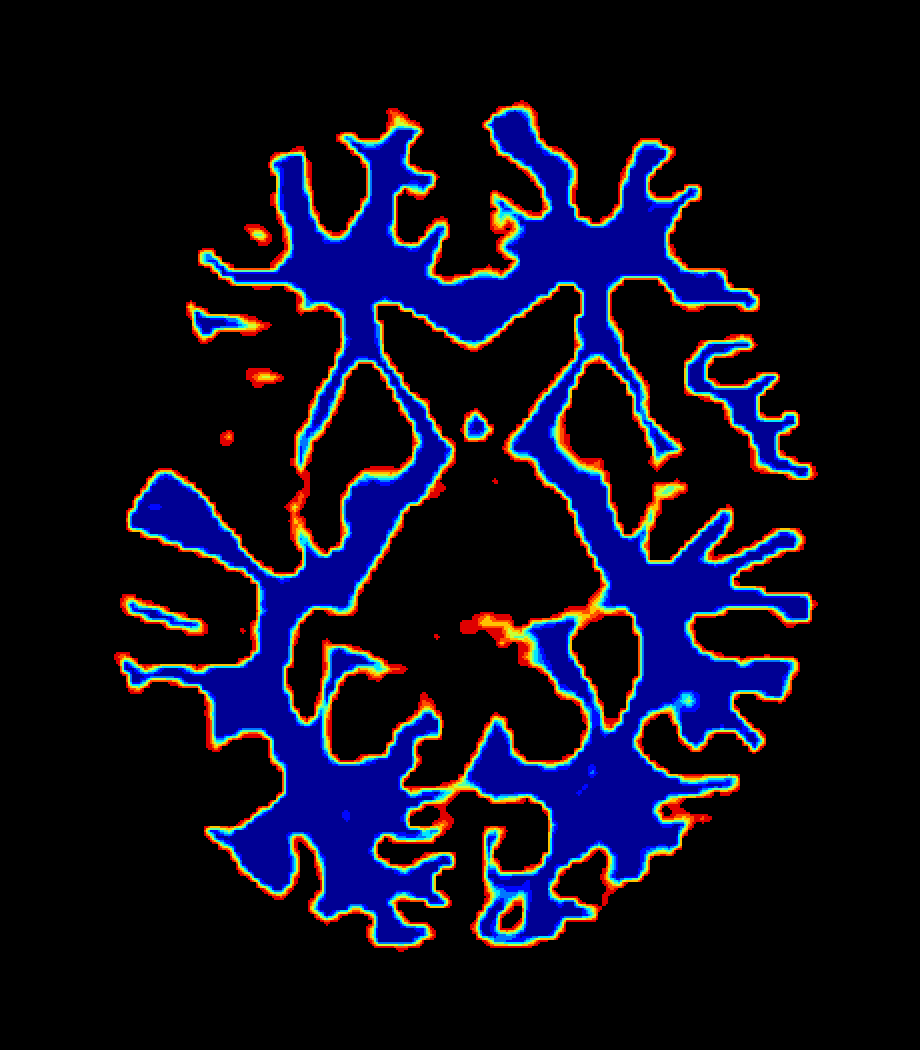}
        }  
\caption{Probability maps obtained for each tissue of two test subjects. The left column of each example depicts the probability maps obtained from the baseline CNN (i.e. single CNN method). The right column shows voxel-wise segmentation confidence of the ensemble of CNNs. Dark blue indicates highest confidence and dark red lowest confidence.}
\label{fig:probMaps}
\end{figure}

\begin{figure}[ht!]
     \begin{center}
     \mbox{
        \includegraphics[width=0.45\linewidth]{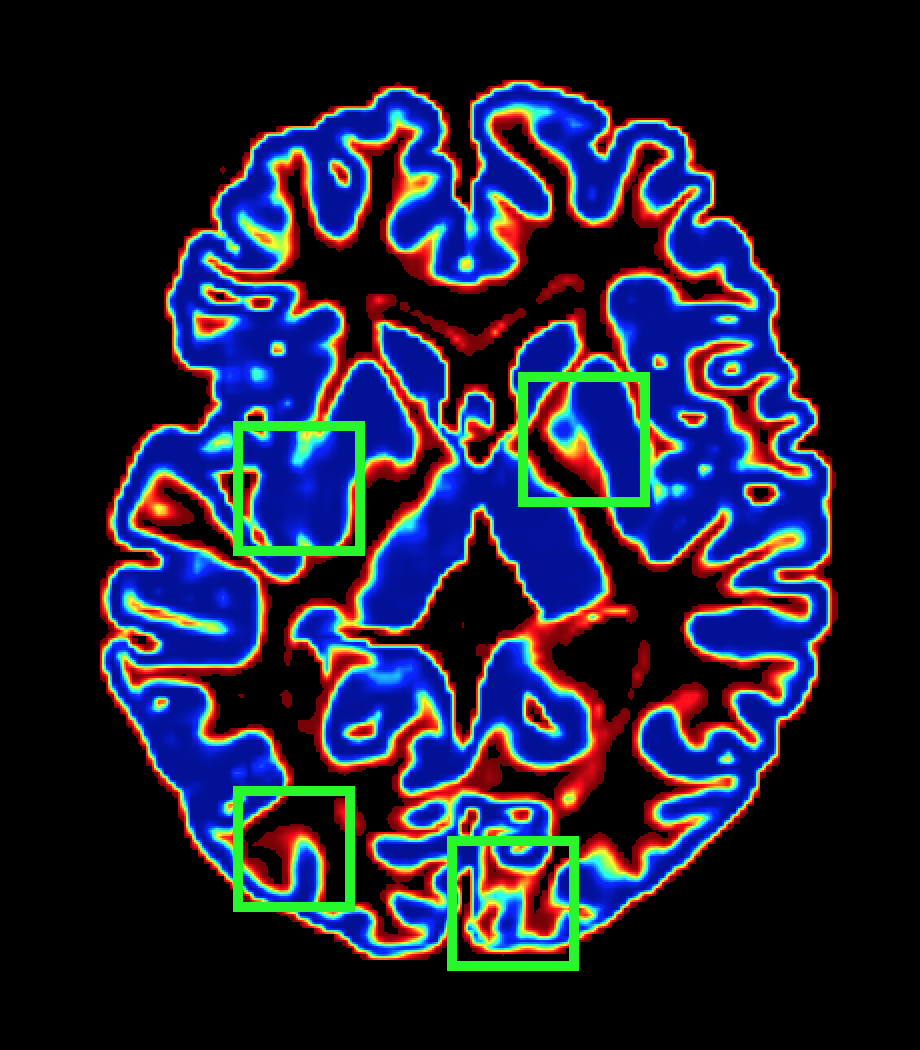}
        \includegraphics[width=0.45\linewidth]{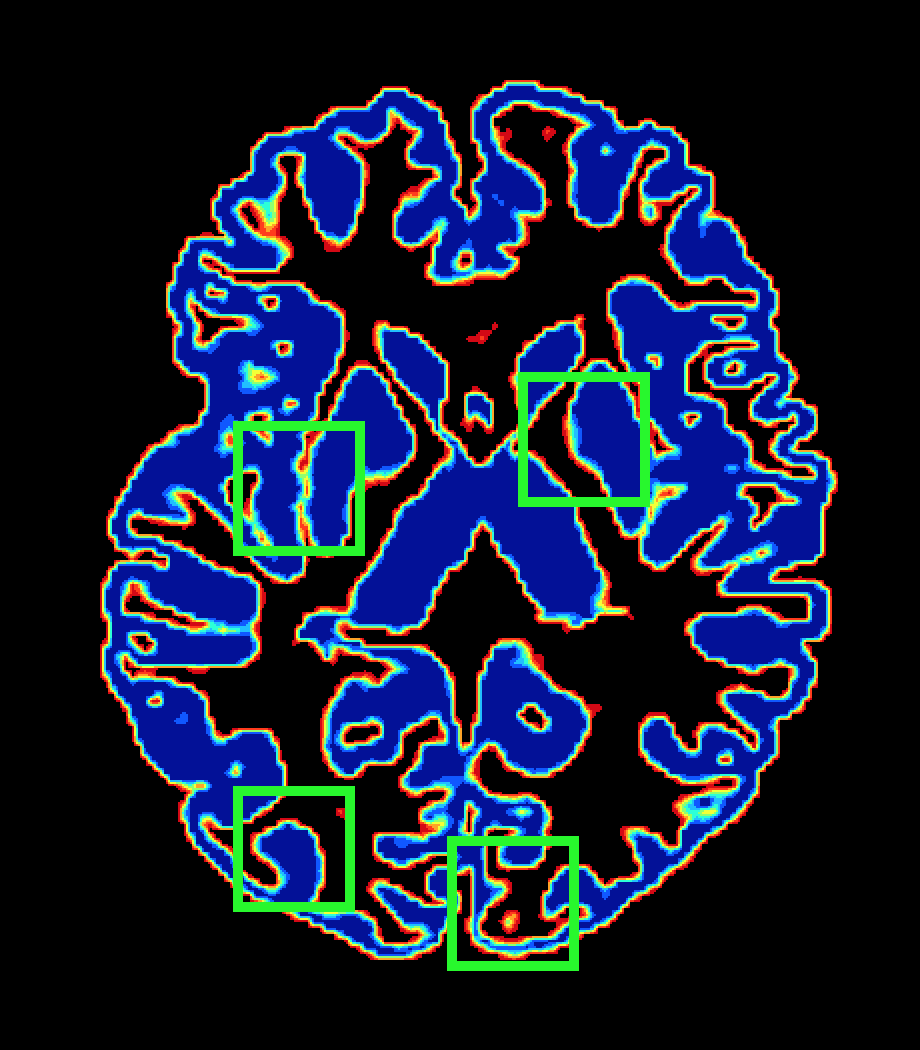}
        }
\caption{Probability maps of gray matter on a 2D slice obtained by a single CNN (\textit{left}) and the ensemble (\textit{right}). Important differences between both methods are highlighted in green.}
\label{fig:probMapsGM}
\end{center}
\end{figure}

To illustrate the impact on reliability of using an ensemble of CNNs, Fig. \ref{fig:probMaps} compares the segmentation confidence of the \textit{Early\_Single} and \textit{Early\_Ensemble} methods, for a given slice of two different subjects. Specifically, the first and third columns depict the probability maps predicted by the single CNN, while the second and fourth columns show the agreement score of the ensemble (i.e., the percentage of CNNs in the ensemble that predicted a given label). We see that the ensemble agreement values are sharper (i.e., closer to 0 -- black or 1 -- blue) than the predictions of the single CNN. With respect to tissue classes, the differences in confidence are smallest for the CSF, and more significant for WM and GM. This is in line with previous results in Table \ref{table:results}, indicating CSF to be an easier tissue to segment. These differences between the methods are illustrated in Fig. \ref{fig:probMapsGM}, showing the ensemble's ability to improve the predictions of a single CNN. 

\subsection{Suggestion of local corrections} 

\begin{figure}[ht!]
     \begin{center}
     \mbox{
        \includegraphics[width=0.95\linewidth]{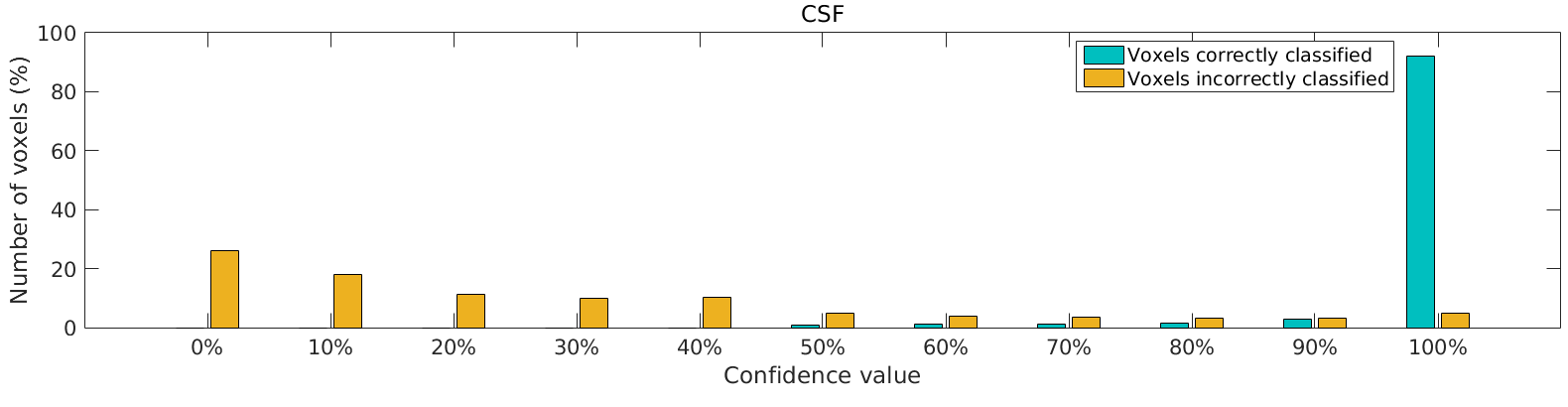}
        }
     \mbox{
        \includegraphics[width=0.95\linewidth]{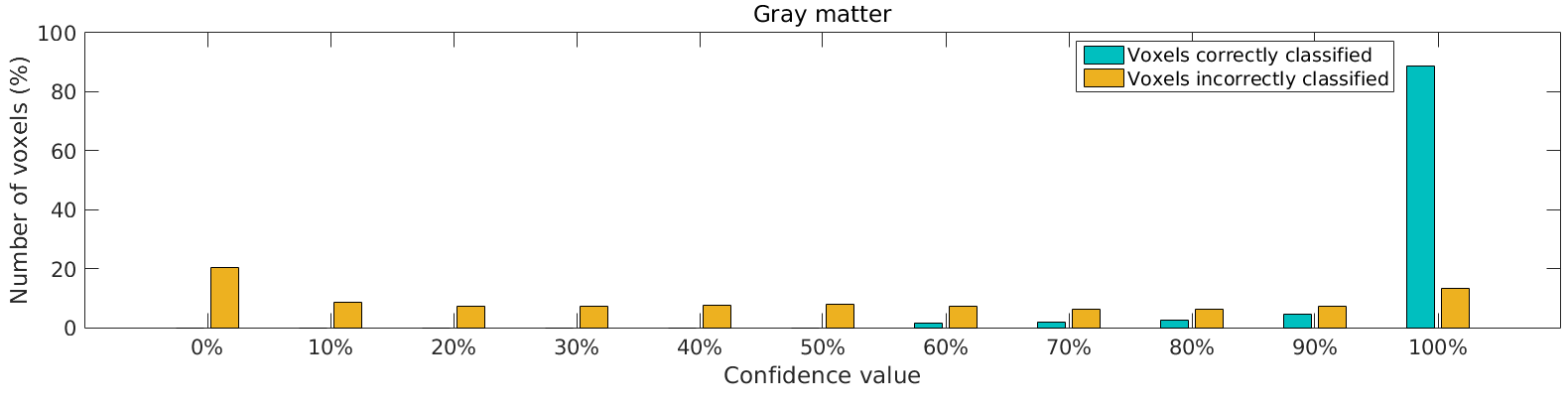}
        }
   \mbox{
        \includegraphics[width=0.95\linewidth]{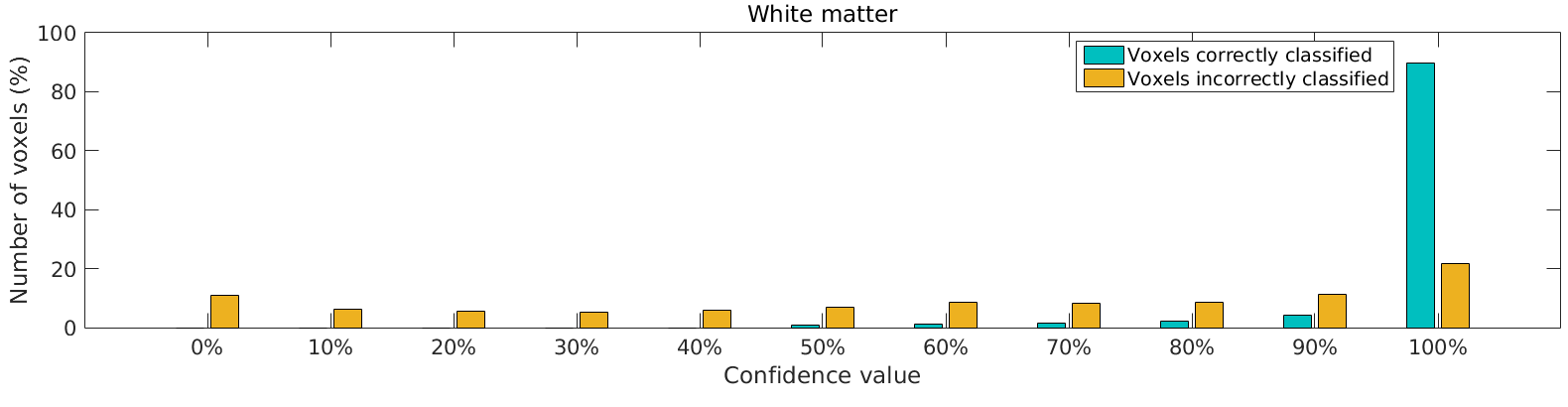}
        }     
\caption{Distribution of confidence values for correctly (\textit{blue bars}) and incorrectly (\textit{yellow bars}) classified voxels, for the three tissues of a given subject.}
\label{fig:PDF_Condifence}
\end{center}
\end{figure}

We validated the usefulness of the ensemble as to suggesting local corrections by comparing confidence values at individual voxels with segmentation errors. Since the ground truth is not available for test instances of the iSEG segmentation challenge, we used a cross-validation strategy for this task. Given the 10 subjects with reference segmentations, we selected a single subject for testing and used the remaining 9 to train the $k=10$ CNNs of the ensemble. For each CNN, 7 subjects were randomly selected for training and the other 2 kept for validation. This process was repeated 5 times, each involving a different test subject. To evaluate the relationship between ensemble confidence and error, we separated confidence values in two groups using an agreement threshold of 60\%:  \emph{low confidence}, where agreement among the ensemble's CNNs equal to or less than 60$\%$ (no more than 6 CNNs agree), and \emph{high confidence}, represented by agreement values greater than 60$\%$ (at least 7 CNNs agree).
Table \ref{tab:correlation} gives the correlation between the \textit{Early\_Ensemble} method's prediction and ground truth value, for low confidence and high confidence voxels. We observe a very high correlation ($>$ 90$\%$) between highly confident predictions and the ground truth, for all three tissues. When prediction is low, correlation drops to values around 50$\%$ for gray matter, and around 10$\%$ for the other two tissues. Furthermore, Fig. \ref{fig:PDF_Condifence} shows the distributions of correctly and incorrectly classified voxels according to their confidence. For all three tissues, most correctly classified voxels have a 100$\%$ agreement, whereas confidence values of incorrectly classified voxels are more evenly distributed. We note that this distribution differs across tissue types. For CSF, voxels classified incorrectly mostly have a low confidence, while more balanced distributions are observed for GM and WM. Again, this indicates the higher difficulty of segmenting these two tissues compared to CSF. Overall, these results validate our hypothesis that the spatial map of ensemble CNN agreement values can be used to suggest local corrections. 


\begin{table}[ht!]
\footnotesize
\centering
\caption{Correlation between the prediction of the proposed method and the ground truth in regions of high and low confidence (values in parenthesis correspond to the percentage of voxels belonging to each of these regions.)}
\label{tab:correlation}
\begin{tabular}{lcccccc}
\toprule
 & \multicolumn{2}{c}{\textbf{CSF}} & \multicolumn{2}{c}{\textbf{GM}} & \multicolumn{2}{c}{\textbf{WM}} \\ 
\cmidrule(lr){2-3}\cmidrule(lr){4-5}\cmidrule(lr){6-7}
& High Conf. & Low Conf. & High Conf. & Low Conf. & High Conf. & Low Conf.  \\ 
\midrule\midrule
Subject\_01 &  0.92 (99.20$\%$) & 0.11 (0.80$\%$)   & 0.93 (96.81$\%$) &  0.48 (3.19$\%$) & 0.92 (97.32$\%$) &  0.10 (2.68$\%$)  \\
Subject\_02 &  0.93 (99.02$\%$) &  0.13 (0.98$\%$)  & 0.94 (96.60$\%$) & 0.51 (3.40$\%$) & 0.94 (97.44$\%$) &  0.10 (2.57$\%$) \\
Subject\_03 &  0.94 (99.12$\%$) & 0.10 (0.88$\%$)  & 0.94 (96.68$\%$)  & 0.49 (3.32$\%$)  & 0.94 (97.43$\%$) & 0.09 (2.57$\%$)   \\
Subject\_04 & 0.91 (99.18$\%$) & 0.07 (0.82$\%$)  & 0.92 (96.79$\%$)   &  0.49 (3.21$\%$)&  0.92 (97.52$\%$) & 0.09 (2.48$\%$)    \\
Subject\_05 & 0.93 (99.05$\%$)   &  0.08 (0.95$\%$)  & 0.94 (96.76$\%$)  & 0.48 (3.24$\%$) & 0.94 (97.59$\%$) &  0.11 (2.41$\%$)                \\ 
\bottomrule
\end{tabular}
\end{table}

\begin{figure}[ht!]
     \begin{center}
     \mbox{
        \includegraphics[width=0.24\linewidth]{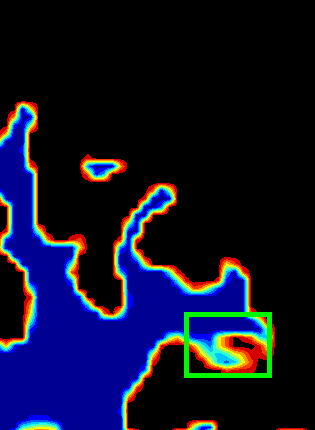}
        \hspace{-2.5 mm}
        \includegraphics[width=0.24\linewidth]{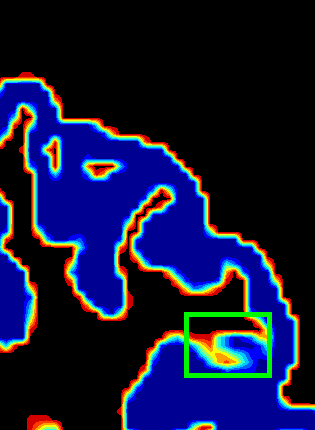}
        \hspace{-2.5 mm}
        \includegraphics[width=0.24\linewidth]{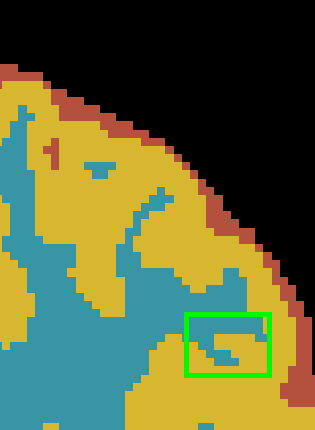} 
        \hspace{-2.5 mm}
        \includegraphics[width=0.24\linewidth]{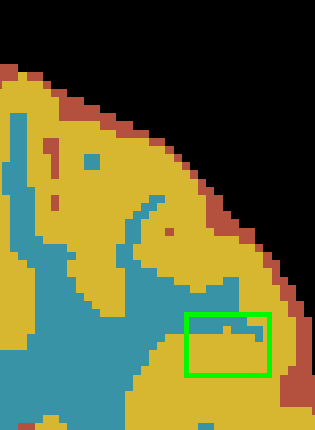}
        }
    \mbox{
        \includegraphics[width=0.24\linewidth]{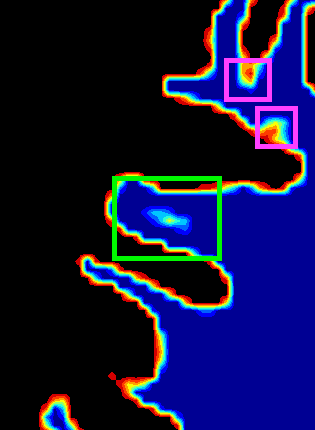}
        \hspace{-2.5 mm}
        \includegraphics[width=0.24\linewidth]{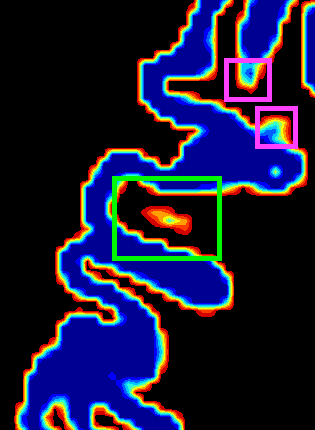}
        \hspace{-2.5 mm}
        \includegraphics[width=0.24\linewidth]{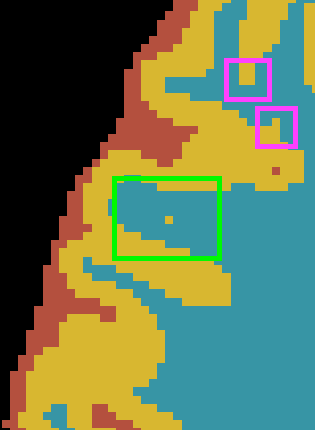}
        \hspace{-2.5 mm}
        \includegraphics[width=0.24\linewidth]{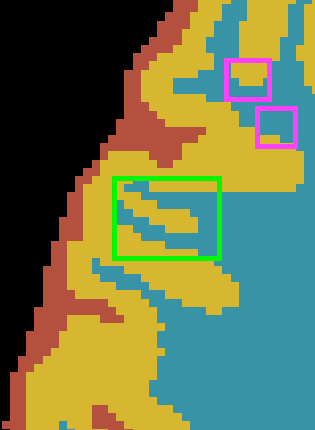}
        }    
\caption{Visual inspection of segmentation confidence values. From left to right: white matter, gray matter, predicted segmentation and reference labels. Green boxes highlight regions with large differences between the predicted contours and the reference standard, where as pink boxes are used to indicate small differences. }
\label{fig:confidence}
\end{center}
\end{figure}

As qualitative validation, Fig. \ref{fig:confidence} shows the examples of confidence values obtained for WM and GM (i.e., percentage of CNNs that predicted these tissues), along with the predicted GT labels. Dark blue corresponds to a total agreement (i.e., 100$\%$ of CNNs voted for the tissue), while dark red indicates the lowest possible agreement (i.e., a single CNN voted for the tissue). Voxels with low confidence (light blue and yellow colors) thus indicate regions with potential segmentation errors, which should be verified by the expert. Visual inspection of these confidence regions, highlighted by the green and pink squares in the figure, corroborates this hypothesis.

\section{Discussion}\label{sec:discussion}

We presented an ensemble-learning approach, which combines multiple deep CNNs to segment isointense infant brain MRI and suggest local corrections in regions of low confidence. In the proposed CNN architecture, multi-modal MR images are employed to deal with the problem of low contrast, using an early or late fusion strategy to combine these different modalities. Furthermore, global and local features are considered in the classification by connecting feature maps from all convolutional layers to the first fully-connected layer. This semi-dense architecture facilitates the propagation of gradients during training, while limiting the number of network parameters.

To improve the generalization of our segmentation approach, we adopted an ensemble-learning strategy that combines the output of 10 CNN models by majority voting. Having several independent predictions also allowed us to build a spatial map of ensemble confidence, which measures segmentation reliability at a voxel level. We showed that confidence values can identify potential errors in the automated segmentation, which might require corrections. The proposed approach was evaluated on the iSEG MICCAI Grand Challenge, and compared to 21 other competing teams. Our approach achieved a state-of-the-art performance, obtaining the highest score in most cases.

The proposed method extends our previous work in \cite{DolzNeuro2017} by considering multi-modal images in the network. An important design choice is the strategy as to merging multiple sources of information. Different modalities can be combined in a tensor and fed to the network. Another option is to create independent paths for the modalities, and fuse the features of these paths at the end of the network, as in Fig. \ref{fig:CNN_archit_Early}. Since satisfactory results have been reported in the literature for both strategies, we investigated in this work the effect of an \textit{early} or \textit{late} fusion. From Table \ref{table:results}, we can see that fusing images in an early stage typically improved the segmentation for most metrics and for all three brain tissues. These findings are not consistent with the results in \cite{nie2016fully}, where combining extracted features in a late stage improved segmentation performance. However, unlike this study, these results were obtained using three image modalities (i.e., T1w,T2w, and FA) instead of two, and used a set of subjects different from the one available for this challenge.

As reported in Table \ref{table:results}, combining predictions from several models in an ensemble yields improvements over a single CNN. This confirms the results obtained in computer vision studies, in the context of color image classification \cite{krizhevsky2012imagenet,zeiler2014visualizing}. These studies showed that ensemble learning can boost the performances of deep neural networks. 
In the context of medical image segmentation, a very valuable benefit of ensemble learning is the ability to measure segmentation confidence at a voxel-level, which can be used to identify regions that need further inspections by medical experts. We validated this assumption by computing the correlation between automatic predictions and manual annotations obtained for 5 subjects, in regions presenting low or high confidence. While high-confidence regions were highly correlated ($>$ 90$\%$) to ground-truth values across all three brain structures, low-confidence regions were poorly correlated, in particular for the CSF and WM ($\approx$ 10$\%$). 


A wide range of techniques were proposed to segment infant brain tissues in MRI (see Table \ref{table:table_RV_seg}), many of them based on atlases. Although atlas-based approaches were successful in segmenting adult brain images, their application to infant brains is more prone to errors due to the poor tissue contrast and the high spatial variability of the infant population. Such approaches also depend on the image registration step, which is time-consuming and a source of errors. In recent studies, deep CNNs were shown to outperform atlas-based methods for this task. For example, Nie et al. \cite{nie2016fully} obtained mean DSC values of 0.852, 0.873 and 0.887 for the CSF, GM and WM, respectively, over 10 subjects. Yet, a limitation of this work was the use of 2D convolutions, which omits important 3D context. Via 3D convolutions, our approach captures spatial information in volumetric data, which is confirmed by a performance improvement over 2D CNN models. Comparing the proposed \textit{Early\_Ensemble} approach to the top-5 ranked methods of the iSEG Grand Challenge, our approach achieved very competitive results, ranking first or second in most cases. A paired sample t-test showed that the difference between our approach and the other best ranked method in the challenge (i.e.,  MSL\_SKKU) is not statistically significant.

Although various CNN-based networks have been used successfully for medical image segmentation (e.g., 3D U-Net \cite{cciccek20163d} or DenseVoxNet \cite{yu2017automatic}), an important benefit of our architecture is the reduced number of trainable parameters. While U-Net and DenseVoxNet require nearly 19M and 4M of parameters, respectively, our semidense network has less than 1M of parameters (nearly 900,000), which results in a 75$\%$-90$\%$ lighter model. This parameter efficiency translates into lower training and segmentation times. For example 3D-UNet requires 3 days for training, whereas our network can be trained in approximately 17 hours using the same hardware. 




A limitation of this study comes from the fact that the ground-truth labels of test instances were not available to the participating teams, which makes interpretation of the results difficult. Based on our cross-validation analysis, we can however conjecture that most errors come from misclassified voxels at the boundaries of GM and WM regions, which have low contrast. Confidence maps show that regions with lowest confidence typically correspond to the borders between these two tissues. Moreover, as mentioned on the iSEG website, manually correcting the data of a single subject took approximately one week. Taking into account that nearly 500 typically developing children will be scanned for the Baby Connectome Project, the adoption of an automatic segmentation tool is highly needed. As reported in our results, the proposed network can segment the data of a subject in 1-2 minutes on a single GPU, or in a few seconds on several GPUs. In addition to its efficiency, the proposed approach can identify potential errors in the automatic segmentation, which could be corrected with limited interactions from the user. A potential extension of this study would be to evaluate our approach in a real clinical setting, in which segmentation time and accuracy is measured for multiple manual raters.



\section{Conclusion}

We presented a novel method based on an ensemble of deep CNNs to segment isointense infant brains in multi-modal MRI images. Our fully-convolutional (FCNN) network considers the 3D spatial context of volumetric data and models both local and global information in the segmentation. We investigated a semi-dense architecture, where the features of each convolutional layer are aggregated as input to the first fully-connected layer. Moreover, two different strategies were investigated to combine multiple input image modalities, using either an early or a late fusion of these modalities. An ensemble learning technique, in which the prediction of 10 CNNs are combined using majority voting, was employed to improve the generalization performance of our method. This ensemble also enables to measure segmentation confidence, using the number of CNNs voting for a particular label. While ensembles  were used in the past to suggest images for annotation, to our knowledge, this is the first work that investigated the problem at a voxel level.

The performance of the proposed method was evaluated in the MICCAI iSEG-2017 Grand Challenge on 6-month infant brain MRI segmentation. The results show that our method is very competitive, ranking first or second among 21 competing teams for most of the metrics. Our experiments also demonstrate the benefit of combining the prediction of multiple CNNs, with performance improvements over using a single CNN, and show the link between ensemble agreement and segmentation error. This suggests that our method has the potential to achieve expert-level performance with limited user interactions. 

\section*{Acknowledgments}

This work is supported by the National Science and Engineering Research Council of Canada (NSERC), discovery grant program, and by the ETS Research Chair on Artificial Intelligence in Medical Imaging.

\section*{References}


\end{document}